\documentclass[12pt]{article}

\usepackage[utf8]{inputenc}
\usepackage[T1]{fontenc}
\usepackage{mathptmx}                 
\usepackage[scaled=0.92]{helvet}      
\usepackage[margin=1in]{geometry}
\usepackage{setspace}
\usepackage{amsmath,amssymb}
\usepackage{graphicx}
\usepackage{tikz}
\usetikzlibrary{positioning,arrows.meta}
\usepackage{booktabs}
\usepackage{array}
\usepackage{multirow}
\usepackage{ragged2e}
\usepackage{enumitem}
\usepackage{caption}
\usepackage{titlesec}
\usepackage[hidelinks]{hyperref}
\newcommand{\fitwidth}[1]{\resizebox{\ifdim\width>\linewidth \linewidth\else\width\fi}{!}{#1}}
\usepackage[natbibapa]{apacite}       
\makeatletter\NAT@longnamesfalse\makeatother
\setcitestyle{aysep={,}}
\usepackage[protrusion=true,expansion=false]{microtype}
\usepackage{changes}
\setaddedmarkup{{\color{blue!65!black}#1}}
\setdeletedmarkup{{\color{red!70!black}#1}}

\titleformat{\section}{\normalfont\bfseries\centering}{\thesection}{0.6em}{\MakeUppercase}
\titleformat{\subsection}{\normalfont\bfseries\centering}{\thesubsection}{0.6em}{}
\titleformat{\subsubsection}{\normalfont\bfseries\raggedright}{\thesubsubsection}{0.6em}{}
\titlespacing*{\section}{0pt}{8pt}{4pt}
\titlespacing*{\subsection}{0pt}{6pt}{3pt}
\titlespacing*{\subsubsection}{0pt}{6pt}{3pt}

\newcommand{\tabfont}{\rmfamily\footnotesize}

\newcommand{\stat}[1]{\noindent\textbf{#1}\ }

\setlist{nosep}

\newif\ifpublic
\publictrue   

\title{\textbf{All Explanations are Wrong, But Many Are Useful:\\
Exploring the Rashomon Explanation Set with Large Language Models}}
\date{}
\ifpublic
  \author{
    Pan Li \\ \normalsize Scheller College of Business, Georgia Tech. \texttt{pan.li@scheller.gatech.edu}}
\else
  \author{}   
\fi

\begin{document}
\doublespacing
\setdisplayskipstretch{0.5}

\begin{singlespace}
\maketitle
\end{singlespace}
\thispagestyle{plain}
\vspace{\ifpublic-1.5em\else-4.5em\fi}

\begin{center}\textbf{Abstract}\end{center}
\vspace{-1.5em}
\begin{singlespace}
Explaining machine-learning models is increasingly important for decision-making and consumer trust, yet it is widely believed to come at a cost: existing Explainable AI (XAI) methods suffer from a persistent accuracy--explainability trade-off. We argue that this trade-off is not fundamental, but rather an artifact of treating explanation and prediction as two separate objectives; when the two are properly coupled, they become complementary, such that equipping a model to explain itself improves, rather than degrades, its predictive accuracy. Realizing this benefit, however, requires overcoming three limitations of existing XAI methods: (1) no single explanation remains valid under distribution shift; (2) scalar attributions, such as SHAP, cannot capture conditional or non-linear relationships; and (3) explanations are disconnected from the predictions they inform. We therefore introduce the \emph{Rashomon Explanation} paradigm for building a set of faithful, prediction-guiding explanations rather than a single one, and we prove that this set is generally non-empty and that the fidelity of these explanations bounds the performance of the models they guide.

To explore the set of Rashomon Explanation, motivated by the organizational theories of sensemaking and double-loop learning, we propose \emph{RashomonLLM}, an Explanation--Prediction--Reflection LLM agentic workflow that generates the explanation set in natural language by iteratively aligning the explanation with the prediction. We prove that under explicit conditions, RashomonLLM converges and recovers the full Rashomon Explanation set. Empirically, across three applications of customer-churn classification, clinical survival regression, and industrial click-through prediction on large-scale logs from Kuaishou's live-streaming platform, RashomonLLM significantly outperforms state-of-the-art prediction and XAI baselines on both accuracy and explanation quality, and surpasses strong specialized deep CTR models. Furthermore, ablation, robustness, and calibration analyses confirm that this accuracy gain is achieved by equipping the model to explain itself and is driven by explanation fidelity, and that it persists across distribution shifts, temporal splits, and random seeds. As a result, our framework enables organizations to advance business performance and lay the groundwork for consumer trust at the same time.

\end{singlespace}

\section{Introduction}\label{sec:intro}

Explainable artificial intelligence (XAI) has become increasingly popular thanks to its benefits in establishing consumer trust \citep{mckinsey2022}, enhancing accountability, model diagnostics, and usability \citep{senoner2021,wangT2022}, even though black-box alternatives (such as Deep Learning or LLMs) may achieve better performance. An example is loan approvals, where financial institutions use a model to assess the risk of an individual defaulting on the payment and generate an ``approve/deny'' decision. In this case, interpretable models\footnote{Some literature (e.g., \citealp{lipton2018}) argues that there are subtle differences between the terminology of ``explanation'' and ``interpretation.'' However, since the majority of the prior work (surveyed in \citep{nauta2023}) does not explicitly distinguish between them, we will use these two terms interchangeably in the rest of this paper.} (such as Logistic Regression or Decision Tree) remain competitive baselines and are widely used in credit scoring \citep{lessmann2015}, since the cost of being wrong and not knowing why is far greater than the benefit of being slightly more right; in fact, the Equal Credit Opportunity Act (ECOA) requires creditors to provide a specific reason if they take an adverse action against a consumer, such as denying their loan application \citep{barocas2016}, and it is obligatory for creditors to disclose these specific reasons to ensure fairness and transparency in the process.

Meanwhile, XAI is a challenging task in practice due to a widely assumed accuracy--explainability trade-off \citep{gunning2019}, in which a gain in interpretability is typically associated with a loss in predictive performance. An organization therefore faces an unappealing choice: deploy a high-performing black box and bolt on a separate post-hoc explainer to approximate its reasoning, or adopt an intrinsically interpretable model and accept what amounts to an ``interpretability tax'' on accuracy. Neither option is satisfactory, as the first yields explanations that are detached from, and may misrepresent, the model that actually makes the decision, while the second forgoes predictive value that the organization could otherwise realize. We argue in this paper that this trade-off is not fundamental, but rather an artifact of treating explanation and prediction as separate objectives. Specifically, we demonstrate that when the two are instead properly coupled, explanation and prediction behave as complements rather than substitutes \citep{milgrom1990}, where equipping a model to explain itself faithfully\footnote{We use \emph{faithfulness} (equivalently, \emph{fidelity}) to refer to the degree to which an explanation reflects the true feature--outcome relationships in the data \citep{jacovi2020,nauta2023}, i.e., its closeness to the ground-truth explanation $E_{S}$ in our Definition~3.} can actually improve its accuracy.

Realizing this benefit based on existing XAI methods, however, is non-trivial, as we establish three major XAI challenges that hinder the reliability and practicality of the produced explanations:

\begin{enumerate}[label=(\alph*),leftmargin=2em]
\item \textbf{No single explanation method can provide valid explanations}, since any explanation that simplifies the underlying model must err on some inputs, and Theorem~1 (presented in Section~\ref{sec:prelim}) shows that a small distributional shift toward such an error region suffices to expose it, so every simplifying explanation fails a suitably chosen ``perturbation test''. This is because with a finite set of observed data, we cannot uniquely determine the true underlying pattern, and any single explanation is just a bet on one of many equally plausible stories, which will eventually go wrong \citep{rudin2019,ragodos2024}.
\item \textbf{Scalar explanations cannot represent complex relationships well}, since scalar weights (e.g., those produced by LIME or SHAP) work best if the underlying relationship is linear; however, if we deal with non-linear or conditional relationships, the value of such scalar explanations will be greatly diminished \citep{kumar2020}: they can only answer ``how much does this feature matter?'', but the question that actually matters is ``how much does this feature matter, for whom, under what conditions, and in combination with what else?'', and this requires a natural-language type of explanation to properly describe these relationships.
\item \textbf{Explanations are disconnected from the prediction process}, since the majority of XAI methods (e.g., post-hoc explanations) operate independently from the prediction model, and do not directly contribute to the prediction performance. However, an explanation that does not help with predictions cannot be verified to reveal the true data patterns, as it is otherwise simply a story that fits the observed data. We show through motivating examples and Theorem~2 that an explanation should be constructed with the goal of revealing the relationships as well as improving predictions, in contrast to various inconsistent goals \citep{han2022}.
\end{enumerate}

\begin{table}[htbp]
\centering
\caption{Summary of Challenges in Existing XAI Methods, Our Solutions, and Validations in Experiments.}
\label{tbl:summary}
{\tabfont
\fitwidth{%
\begin{tabular}{>{\RaggedRight\arraybackslash}m{0.26\linewidth} >{\RaggedRight\arraybackslash}m{0.27\linewidth} >{\centering\arraybackslash}m{0.15\linewidth} >{\RaggedRight\arraybackslash}m{0.22\linewidth}}
\toprule
\textbf{Existing Challenges} & \textbf{Our Proposed Solutions} & \textbf{Theoretical Validations} & \textbf{Empirical Validations}\\
\midrule
\#1: No single method can provide valid explanations & Construct a set of explanations as Rashomon Explanation & Theorem~1 & Comparison with single-explanation baselines\\
\addlinespace
\#2: Scalars cannot represent complex relationships well & An LLM agentic workflow for producing explanations & Definition~3; Assumption~4 & Comparison with scalar-representation baselines\\
\addlinespace
\#3: Explanations are disconnected from predictions & A novel framework for aligning explanation and prediction & Theorem~2 & Ablation studies with prediction-only models\\
\bottomrule
\end{tabular}}}
\end{table}

To address these three challenges, we propose a novel explanation paradigm (``\emph{Rashomon Explanation}'') and a specific explanation method (``\emph{RashomonLLM}''), and we provide both theoretical and empirical validations of them in this paper, which are summarized in Table~\ref{tbl:summary}:

First, instead of relying on a single explanation model, we advocate for constructing a set of explanations, which we term ``Rashomon Explanation'' in Section~\ref{sec:prelim}, that have similar levels of fidelity and perform equally well when utilized by prediction models. The final explanation is produced as the aggregation of the entire Rashomon Explanation set. This concept enables us to utilize the ``collective intelligence'' of a set of explanations to produce more robust, reliable, and stable explanations, which we demonstrate both theoretically and empirically in this paper.

Second, to generate natural language explanations, we select the powerful Large Language Model (LLM) as the backbone functional class of our method, since its strong reasoning capability makes it effective at capturing non-linear, conditional relationships and multivariate interactions prevalent in real-world applications \citep{kojima2022}. It also enables us to explore the Rashomon Explanation set with theoretical guarantees, as we establish in Theorem~4.

Third, we propose an agentic framework of ``Explanation--Prediction--Reflection'' (EPR) with three LLM agents: an Explanation Agent, which produces explanations based on predictions; a Prediction Agent, which produces predictions based on provided explanations; and a Reflection Agent, which updates the prompting strategy to better align explanations and predictions. These three agents are repeatedly optimized following an alternating training procedure until the prediction error stabilizes, and the explanations stop updating. Our framework instantiates double-loop learning \citep{argyris1997} to revise how it explains, while tying each explanation to the prediction it serves, reflecting the action-guiding criterion of sensemaking \citep{weick1995}. This enables us to obtain more stable and robust explanations, as well as better model performance.

Putting all these three elements together, we address the foundational challenge in XAI methods: while no single explanation model can provide full and valid explanations (i.e., ``\emph{All Explanations are Wrong}''), a collection of many explanations (Rashomon Explanation) has a high explanation fidelity and helps increase the performance of the focal task when included as model inputs (i.e., ``\emph{But Many Are Useful}''). We validate the benefits of RashomonLLM with extensive theoretical analysis, as well as empirical evaluations on two public Kaggle benchmarks and large-scale interaction logs from Kuaishou's live-streaming platform. We show that RashomonLLM consistently generates explanations with significantly better quality and produces significantly better prediction performance, compared to state-of-the-art prediction models, interpretable machine learning models, and deep CTR models. We conduct additional analyses to show that the prediction gains are driven by the fidelity of the explanations. In sum, our model works well with a wide range of tabular data, and with our findings, we hope that practitioners will further embrace XAI.

In this paper, we make the following contributions. First, we establish, theoretically and empirically, that explanation and prediction are \emph{complements rather than substitutes}, as equipping a model to explain itself faithfully does not tax its accuracy but \emph{improves} it. Second, to make this complementarity attainable, we introduce the Rashomon Explanation paradigm representing a set of faithful, prediction-guiding explanations, and propose RashomonLLM, an agentic workflow that constructs this set.\footnote{Our implementation of RashomonLLM, together with the scripts and data-processing pipeline used in our experiments, is available at the anonymized repository \url{https://anonymous.4open.science/r/Rashomon_LLM-C283}.} Third, we develop the formal guarantees (Theorems~1--4) to demonstrate the validity of our paradigm and the advantages of our proposed model. Finally, we conduct extensive experiments to illustrate the empirical benefits of RashomonLLM, as it significantly outperforms state-of-the-art baselines across a wide range of experiment settings, and is capable of advancing business performance and laying the groundwork for consumer trust at the same time.

\section{Related Work}\label{sec:relwork}

\subsection{Benefits of Explanations, and Explainable Machine Learning Methods}\label{sec:rw-benefits}

The study of system-generated explanations has a long tradition in IS \citep{gregor1999}, as a stream of research has examined how explanations affect user trust and decision quality in recommendations \citep{wang2007,xu2014}. A report \citep{mckinsey2022} shows that companies making their AI explainable are more likely to see their revenues grow by 10\% or more. Accordingly, XAI methods have been widely implemented by stakeholders to better understand decision-making \citep{chenZ2022}, increase consumer engagement \citep{liY2020,wangT2022}, and provide better guidance for product design \citep{zhang2021}. 

We categorize existing XAI methods into three groups \citep{rudin2019,nauta2023}:

\begin{enumerate}[label=(\arabic*),leftmargin=2em]
\item \textbf{Post-hoc Methods}, such as LIME \citep{ribeiro2016}, SHAP \citep{lundberg2017}, and ROLEX \citep{kim2023}, which produce explanations based on observed outcomes. They utilize feature attribution \citep{sundararajan2017}, propagation \citep{bach2015}, or model inspection \citep{bau2017} techniques to explain model outputs in a post-hoc manner.
\item \textbf{Built-in Methods}, such as the attention mechanism \citep{vaswani2017} and prototype learning \citep{chenC2019}, which utilize the nature of intrinsically interpretable machine learning models to produce explanations directly (attention values, prototypes, etc.).
\item \textbf{Supervised Methods}, which utilize ground-truth explanations provided a priori to train the supervised model to produce both explanations and prediction outcomes \citep{plumb2018}.
\end{enumerate}

While these methods are all useful and widely adopted, there are significant gaps in each approach as summarized in Table~\ref{tbl:xai}, resulting in the accuracy--explainability trade-off \citep{bell2022}. Specifically, post-hoc methods may lead to false characterizations \citep{laugel2019,rudin2019,lakkaraju2020} or unnecessary authority to the black box \citep{rudinradin2019,ragodos2024}, while built-in methods typically experience significant performance loss \citep{lundberg2017,bell2022}. For supervised methods, ground-truth explanations are expensive to obtain in many scenarios, making them not widely applicable \citep{rudin2022,nauta2023}. We tackle these challenges with a novel explanation paradigm in this paper.

\begin{table}[htbp]
\centering
\caption{Summary of Challenges of Existing XAI Methods and Benefits of Our Method.}
\label{tbl:xai}
{\tabfont
\fitwidth{%
\begin{tabular}{>{\RaggedRight}p{0.24\linewidth} >{\RaggedRight}p{0.36\linewidth} >{\RaggedRight\arraybackslash}p{0.36\linewidth}}
\toprule
\textbf{XAI Methods} & \textbf{Problems} & \textbf{Benefits of Our Method}\\
\midrule
Post-hoc Methods & False and unreliable characterizations & Reliable and high-quality explanations\\
Built-in Methods & May sacrifice accuracy & Improve prediction accuracy\\
Supervised Methods & Require ground truth a priori & Do not need ground truth\\
LLM-Based Methods & Hallucination and instability & Stable and high-fidelity\\
\bottomrule
\end{tabular}}}
\end{table}

\subsection{Large Language Models for Producing Explanations}\label{sec:rw-llm}

LLMs become an emerging topic with three unique advantages: (1) \emph{Ease of Use}: No domain expertise is needed when deploying an LLM \citep{jin2024}; (2) \emph{Performance}: LLMs outperform deep learning models across a wide range of tasks, and can rival humans or even experts \citep{schoenegger2025}; (3) \emph{Flexible Thinking}: LLMs can think outside the box and not be constrained by pre-determined rules. In particular, LLMs have great potential to act as post-hoc explainers \citep{kroeger2023,ajwani2024} similar to SHAP, with direct prompting or self-reflection \citep{bilal2025}. In two examples, a prompting method was proposed \citep{wangL2023} to perform visualization recommendations and return human-like explanations at the same time, while LLMs' reasoning capability is leveraged to mine review texts to generate explanations \citep{tsai2024}.

While this is a promising direction, a series of problems surface as shown in Table~\ref{tbl:xai}. In particular, they suffer from hallucination \citep{huangL2025}, since they might generate outputs that are inaccurate or fabricated. In addition, the outputs are typically context-sensitive and unstable \citep[T.][]{brownT2020,marjanovic2024}, where different prompts under different contexts may lead to dramatically different outputs. Under certain scenarios, LLMs are less effective than locally trained machine learning models \citep[K.~E.][]{brownK2025}, and researchers caution against the use of LLM outputs to directly serve the purpose of explanations \citep{barez2025}. We tackle these challenges by proposing an LLM agentic workflow to iteratively align predictions and explanations, leading to significant improvements for both tasks and resolving the instability issue.

\subsection{Research Gaps in the Literature}\label{sec:rw-gaps}

Beyond limitations already summarized in Table~\ref{tbl:xai}, we identify two deeper, behaviorally grounded gaps, each of which provides a direct motivation for our solution.

First, previous methods share the same goal of accurate representation, i.e., faithfully recovering the true relationship between features and model outcomes. However, the Theory of Sensemaking \citep{weick1995} argues that the value of an interpretive account lies not only in its correspondence to an external reality but also in its capacity to guide effective action, which requires a criterion of plausibility and usefulness. From this perspective, the problems of existing XAI methods reflect the misalignment between the representational goal these methods pursue and the pragmatic, action-guiding goal that decision-makers actually require. This misalignment becomes the foundation for our proposed model design to align interpretation results with prediction outcomes.

Second, existing XAI methods share a structural limitation, as they are designed as single-loop systems \citep{argyris1997}, producing explanations within a fixed framework without adjusting in response to observed failures. This single-loop design is particularly consequential for hallucination \citep{huangL2025}, and reflects a systematic misspecification in the explanation paradigm that requires a framework-level revision. We tackle this challenge directly through the design of our iterative agentic workflow that enables a mutually beneficial learning loop in our framework, as complementarity theory suggests \citep{milgrom1990}.

To address these research gaps, we derive a set of design principles in Section~\ref{tbl:theory} that motivate our proposed concept and model architecture to align explainability with decision-making needs.

\section{Preliminaries and Theoretical Motivation}\label{sec:prelim}

In this section, we will introduce the preliminaries and develop the theoretical foundations.

\subsection{``All Explanations are Wrong'' --- The Pitfall of a Single Explanation Model}\label{sec:wrong}

Modern machine learning models are big and complex, and the faithful explanations for them are usually highly complicated and cannot be represented in a trivial manner. In fact, as a general principle in computational complexity \citep{papadimitriou2003}, any explanation simpler than a model must differ from it on some inputs, and it typically involves a loss of information.

In this section, we make an even stronger theoretical claim that no single explanation model that simplifies the underlying model can remain faithful under every distribution. To illustrate this idea, we first consider a simple synthetic example adapted from the literature \citep{ragodos2024}, where we generate 10,000 simulated credit scores using the following formula:
\begin{equation}\label{eq:credit-score}
\text{Credit Score} = 850 + 2.5\times\text{History} - 15\times\text{Inquiries} - 4\times\text{Delinquencies} - 100\times\text{Utilization},
\end{equation}
where the four features are drawn independently from uniform distributions: $\text{History}\sim U[0,50]$, $\text{Inquiries}\sim U[0,10]$, $\text{Delinquencies}\sim U[0,24]$, and $\text{Utilization}\sim U[0,1]$. We then train a series of black-box prediction models that all achieve low error and use SHAP \citep{lundberg2017} to produce model explanations. As summarized in Table~\ref{tbl:shapley}, the resulting explanations differ across these models that are equally accurate, and even the order of relative importance can disagree. The point is not that any single method is uniquely at fault, but that, given only finite observations, we cannot determine from the data alone which one reflects the truth, due to underspecification \citep{damour2022}. Since even this extremely simple, linear case already exhibits the divergence, it can only grow for the more complex, non-linear relationships encountered in practice.

\begin{table}[htbp]
\centering
\caption{Comparison of Ground-Truth and Model-Inferred Shapley Values.}
\label{tbl:shapley}
{\tabfont
\fitwidth{%
\begin{tabular}{lcccc}
\toprule
\textbf{Models} & \textbf{History} & \textbf{Inquiries} & \textbf{Delinquencies} & \textbf{Utilization}\\
\midrule
Ground Truth & 31.25 & 37.5 & 24 & 25\\
\midrule
Random Forest & 29.6 & 35.8 & 22.4 & 24\\
Support Vector Machine & 27.7 & 35.6 & 22.9 & 22.7\\
Multi-Layer Perceptron & 30.5 & 36.5 & 23.4 & 23.8\\
CNN & 35.4 & 37.8 & 25.4 & 22.6\\
RNN & 31.4 & 41.8 & 22.1 & 23.7\\
\bottomrule
\end{tabular}}}
\end{table}

We now present the theoretical result behind the observed pitfalls, where we can always construct an ``adversarial'' set for every single explanation model where it will fail on that set.

\stat{Theorem~1 (Sensitivity of Explanations to Distribution Shift).} Let the loss be bounded, $\ell(\cdot,\cdot) \in [0,B]$. Let $f_{W}$ denote the ground-truth model and $f_{E}$ the explanation-induced model, and suppose they achieve similar loss on the training distribution $P$, $\left| L\left( f_{E},P \right) - L\left( f_{W},P \right) \right| \leq \epsilon_{0}$ for some small $\epsilon_{0} > 0$. Define the pointwise excess loss $g(x,y) = \ell\left( y,f_{E}(x) \right) - \ell\left( y,f_{W}(x) \right)$, and suppose there is a region $A$ on which the explanation is systematically wrong, i.e., $\mathbb{E}_{P}\!\left[ g \mid A \right] \geq \delta > 0$. For any mixing weight $\lambda \in (0,1]$, define the shifted distribution $\widetilde{P} := (1-\lambda)\,P + \lambda\,P(\cdot \mid A)$. Then, the total variation distance $\mathrm{TV}\left( \widetilde{P},P \right) \leq \lambda$ and $\left| L\left( f_{W},\widetilde{P} \right) - L\left( f_{W},P \right) \right| \leq \lambda B$, while
\begin{equation}\label{eq:thm1-gap}
L\left( f_{E},\widetilde{P} \right) - L\left( f_{W},\widetilde{P} \right) \geq \lambda\delta - \epsilon_{0}.
\end{equation}
For any $\Delta \in (0,\ \delta - \epsilon_{0}]$, setting $\lambda = (\Delta + \epsilon_{0})/\delta \in (0,1]$ yields an adversarial distribution on which the ground-truth model remains robust while the explanation model's excess loss is at least $\Delta$.

\stat{Proof.} Since $\widetilde{P} - P = \lambda\left( P(\cdot \mid A) - P \right)$, we have $\mathrm{TV}\left( \widetilde{P},P \right) = \lambda\,\mathrm{TV}\left( P(\cdot \mid A),P \right) \leq \lambda$. For the robustness bound, $L\left( f_{W},\widetilde{P} \right) - L\left( f_{W},P \right) = \lambda\!\left( \mathbb{E}_{P(\cdot \mid A)}\!\left[ \ell_{W} \right] - \mathbb{E}_{P}\!\left[ \ell_{W} \right] \right)$, whose absolute value is at most $\lambda B$ because $\ell \in [0,B]$. Finally, the excess-loss gap is the expectation of $g$ under $\widetilde{P}$, $\mathbb{E}_{\widetilde{P}}[g] = (1-\lambda)\,\mathbb{E}_{P}[g] + \lambda\,\mathbb{E}_{P}\!\left[ g \mid A \right] \geq -(1-\lambda)\epsilon_{0} + \lambda\delta \geq \lambda\delta - \epsilon_{0}$, where we used $\mathbb{E}_{P}[g] \geq -\epsilon_{0}$ (from the training-loss similarity) and $\mathbb{E}_{P}[g \mid A] \geq \delta$. Setting $\lambda\delta - \epsilon_{0} = \Delta$, which requires $\Delta \leq \delta - \epsilon_{0}$ (so that $\lambda = (\Delta+\epsilon_{0})/\delta \in (0,1]$, presuming $\epsilon_{0} < \delta$), gives the stated bound. \hfill Q.E.D.

\stat{Theorem~1 In Plain Language:} a modest distribution shift that re-weights toward a low-probability region where the explanation is wrong leaves the ground-truth model unchanged but exposes the explanation model's hidden error. Any single explanation is therefore a bet on one of many equally plausible stories, and a small shift suffices to make that bet go wrong.

Based on Theorem~1, a single explanation method can only work if the observed data characterizes the entire population, which is impossible with finite data due to underspecification. Therefore, we advocate exploring a collection of explanation methods instead to obtain more stable and robust explanation results, and will present its formal definition next.

\subsection{``But Many Are Useful'' --- How Rashomon Explanation Tackles the Challenge}\label{sec:useful}

We now illustrate the other side of the story, i.e., a collection of many explanations can still be beneficial for the learning task. In particular, we identify the set of explanations, which is sufficiently close to the ground truth and leads to near-optimal model performance, as the ``\emph{Rashomon Explanation}'' in Definition~3. This concept is analogous to both the Rashomon Set \citep{fisher2019} in Definition~1, which represents a set of near-optimal models for a learning task; and the Rashomon Ratio \citep{semenova2022} in Definition~2, which measures the size of those models.

\stat{Definition~1 (Rashomon Set).} Given $\theta \geq 0$, a dataset $S$ drawn from a distribution $P$, a functional space $F$, and the risk $L(f) := \mathbb{E}_{P}[\ell(y,f(x))]$,\footnote{We write $L(f,P')$ for the risk under another distribution $P'$ (e.g., the shifted distribution $\widetilde{P}$ in Theorem~1).} the Rashomon set $R_{set}(F,\ \theta)$ is defined as $R_{set}(F,\ \theta) := \{ f \in F : L(f) \leq L\left( f^{*} \right) + \theta\}$, where $f^{*} = \underset{f \in F}{\arg\min}\,{L(f)}$ is the risk minimizer over $F$.

\stat{Definition~2 (Rashomon Ratio).} The Rashomon ratio $R_{ratio}(R_{set}(F,\ \theta))$ is the ratio of the volume of the set of accurate models to the volume of the functional space $F$. Given a prior $\rho$ on $F$, the general form of Rashomon ratio is $R_{ratio}\left( R_{set}(F,\ \theta) \right) = \int_{f \in F} I(f \in R_{set}(F,\ \theta))\,d\rho(f)$.

\stat{Definition~3 (Rashomon Explanation).} The Rashomon Explanation $R_{explanation}(F,\theta)$ is a set of generated explanations $E$ that is sufficiently close to the ground-truth explanation $E_{S}$, and that each explanation will lead to a near-optimal model $f_{E}$ when it is included as part of model inputs:
\begin{equation}\label{eq:rashomon-explanation}
R_{explanation}(F,\theta) := \{ E : d\left( E,E_{S} \right) \leq \theta;\ f_{E} \in R_{set}(F,\theta)\}
\end{equation}
where $d\left( E,E_{S} \right)$ is a fidelity-distance metric measuring the discrepancy between $E$ and $E_{S}$.

According to this definition, the Rashomon Explanation contains those explanations that are both \textbf{faithful} (within distance $\theta$ of the ground-truth explanation $E_{S}$) and \textbf{useful} (supporting a near-optimal prediction model). Here useful denotes a population-level property, i.e., an explanation is useful when it supports accurate prediction of the underlying pattern, and it is deliberately distinct from a human's \emph{perceived usefulness} of the explanation, a separate behavioral dimension that we leave to future work. We now provide both motivational examples and theoretical analysis to explain why the ``usefulness'' is an important dimension in our definition. To start with, we utilize the same synthetic dataset built in Section~\ref{sec:wrong} and implement a GPT-4o-based prediction model with five prompting strategies: (1) without explanation; (2) with explanation and 10\% noise; (3) with explanation and 5\% noise; (4) with explanation and 1\% noise; (5) with ground-truth explanation. The detailed prompts are outlined in Section~\ref{sec:prompts}.

\begin{table}[htbp]
\centering
\caption{Comparison of LLM Prediction Performance with Different Prompt Design.}
\label{tbl:prompt}
{\tabfont
\fitwidth{%
\begin{tabular}{lcc}
\toprule
\textbf{Prompt Design} & \textbf{RMSE} & \textbf{MAE}\\
\midrule
No Explanation & 5.97 & 4.49\\
Explanation w/ 10\% Noise & 3.36 & 1.78\\
Explanation w/ 5\% Noise & 1.84 & 1.13\\
Explanation w/ 1\% Noise & 0.98 & 0.71\\
Ground-Truth Explanation & 0.81 & 0.63\\
\bottomrule
\end{tabular}}}
\end{table}

Results in Table~\ref{tbl:prompt} demonstrate that even with an imperfect explanation (i.e., with 10\% noise), the LLM will still be able to produce significantly better performance compared to the case when no explanations are provided, and that the performance will become even better when the explanation quality gets better. This observation matches Theorem~2 that we provide below, which establishes the benefits of identifying better explanations for improving the model performance.

\stat{Theorem~2 (Fidelity Bounds Excess Risk).} An explanation $E$ induces an \emph{assumed data model} $Q_{E}$, the distribution against which the explanation-guided model is implicitly optimized. Let the loss be bounded, $\ell(\cdot,\cdot) \in [0,B]$, let $L_{Q_{E}}(f) := \mathbb{E}_{Q_{E}}[\ell(y,f(x))]$ be the risk under $Q_{E}$, and suppose fidelity controls the induced distribution: $D_{\mathrm{KL}}\!\left( P \,\Vert\, Q_{E} \right) \leq \varphi\!\left( d(E,E_{S}) \right)$ for some increasing, continuous $\varphi$ with $\varphi(0) = 0$. Then the model $f_{Q_{E}}^{*} = \arg\min_{f \in F}L_{Q_{E}}(f)$ optimized against $Q_{E}$ satisfies
\begin{equation}\label{eq:thm2-risk}
L\!\left( f_{Q_{E}}^{*} \right) - L\!\left( f^{*} \right) \;\leq\; 2B\sqrt{\tfrac{1}{2}\,D_{\mathrm{KL}}\!\left( P \,\Vert\, Q_{E} \right)} \;\leq\; 2B\sqrt{\tfrac{1}{2}\,\varphi\!\left( d(E,E_{S}) \right)},
\end{equation}
where $f^{*} = \arg\min_{f \in F}L(f)$. In particular, the excess risk vanishes as $d(E,E_{S}) \rightarrow 0$.

\stat{Proof.} For any $f$, since $\ell \in [0,B]$, the difference in risk under the two distributions is bounded: $\left| L(f) - L_{Q_{E}}(f) \right| \leq B\,\mathrm{TV}\!\left( P,Q_{E} \right)$. Therefore $L\!\left( f_{Q_{E}}^{*} \right) \leq L_{Q_{E}}\!\left( f_{Q_{E}}^{*} \right) + B\,\mathrm{TV} \leq L_{Q_{E}}\!\left( f^{*} \right) + B\,\mathrm{TV} \leq L\!\left( f^{*} \right) + 2B\,\mathrm{TV}$, since $f_{Q_{E}}^{*}$ minimizes $L_{Q_{E}}$. Applying Pinsker's inequality, $\mathrm{TV}\!\left( P,Q_{E} \right) \leq \sqrt{\tfrac{1}{2}D_{\mathrm{KL}}(P \Vert Q_{E})}$, and then the fidelity hypothesis yields the stated bound. \hfill Q.E.D.

\stat{Theorem~2 In Plain Language:} The gap between the explanation-guided and optimal models is controlled by the explanation fidelity, and shrinks to zero when approaching the ground truth.

Theorems~1 and~2 establish the two halves of a coupling: no single explanation is robust everywhere (Theorem~1), yet fidelity leads to predictive usefulness (Theorem~2), which motivates our proposed concept. We now show in Theorem~3 that the Rashomon Explanation set exists under our Definition~3, which rests on the following two assumptions.

\stat{Assumption~1 (Explanation-induced class).} The candidate models are built from an explanation, so the relevant functional space is the explanation-induced class $F = \{ f_{E} : E \in \mathcal{E} \}$, and the exploration mechanism draws a finite collection $\mathcal{F} \subseteq F$ of such models i.i.d.\ from a prior $\rho$ on $F$.

\stat{Assumption~2 (Inverse fidelity coupling).} The converse of Theorem~2 holds at least weakly: there exists an increasing function $\psi$ with $\psi(0) = 0$ such that the excess risk is lower-bounded by the infidelity, $R(f_{E}) := L(f_{E}) - L(f^{*}) \geq \psi\!\left( d(E,E_{S}) \right)$; equivalently, a badly unfaithful explanation cannot, by accident, induce a near-optimal model.

\stat{Theorem~3 (Existence of the Rashomon Explanation Set).} Let $F$ be the explanation-induced class with risk minimizer $f^{*} = \arg\min_{f \in F} L(f)$. Suppose $R_{ratio}\left( R_{set}\left( F,\theta_{0} \right) \right) \geq 1 - \varepsilon^{1/|\mathcal{F}|}$, where $|\mathcal{F}|$ is the cardinality of $\mathcal{F}$. Then with probability at least $1 - 2\varepsilon$ there exists $f_{E} \in \mathcal{F}$ with
\begin{equation}\label{eq:thm3-bound}
L(f_{E}) \leq L\left( f^{*} \right) + \theta_{0} + B\sqrt{\frac{\log(1/\varepsilon)}{2n}},
\end{equation}
so that its excess risk satisfies $R(f_{E}) \leq \theta := \theta_{0} + B\sqrt{\log(1/\varepsilon)/(2n)}$, the effective Rashomon radius, where $n$ is the training size and $B$ bounds the loss. Consequently, under Assumption~2, the explanation satisfies $d(E,E_{S}) \leq \psi^{-1}(\theta)$. Therefore, whenever the effective Rashomon radius obeys $\theta \geq \psi^{-1}(\theta)$, the explanation $E$ meets both clauses of Definition~3, and hence $R_{explanation}(F,\theta) \neq \emptyset$.

\stat{Proof.} Existence of the near-optimal model $f_{E} \in \mathcal{F}$ and the loss bound follow from Theorems~5 and~6 of \citet{semenova2022}, applied to the explanation-induced class $F$ of Assumption~1 with the Rashomon ratio taken under the sampling distribution $\rho$; the upper loss bound gives $L(f_{E}) \leq L(f^{*}) + \theta$, i.e., $R(f_{E}) \leq \theta$. The stated probability $1-2\varepsilon$ is over the i.i.d.\ draw of the collection $\mathcal{F}$ from $\rho$ and the training sample of size $n$: the event that no draw lands in $R_{set}(F,\theta_{0})$ and the finite-sample deviation event each have probability at most $\varepsilon$, and combine by a union bound. By Assumption~1 this model is $f_{E}$ for some explanation $E$. By Assumption~2, $\psi\!\left( d(E,E_{S}) \right) \leq R(f_{E}) \leq \theta$, so $d(E,E_{S}) \leq \psi^{-1}(\theta)$. The usefulness clause $f_{E} \in R_{set}(F,\theta)$ then holds since $R(f_{E}) \leq \theta$, and the fidelity clause $d(E,E_{S}) \leq \theta$ holds whenever $\theta \geq \psi^{-1}(\theta)$. Both clauses of Definition~3 are then satisfied, so $E \in R_{explanation}(F,\theta)$. \hfill Q.E.D.

\stat{Theorem~3 In Plain Language:} the existence of a near-optimal model guarantees a faithful and useful explanation, i.e., a member of the Rashomon Explanation set.

We would like to point out that this result not only guarantees the existence, but also illustrates the potential of the Rashomon Explanation set in tackling the challenges in existing XAI methods. Before presenting our proposed method to explore the Rashomon Explanation set, we now summarize the scope of our paper regarding different types of explanations.

\subsection{Categorization of Explanations and the Scope of Our Research}\label{sec:categorization}

The form of an explanation can be categorized along the following four dimensions \citep{linardatos2020,rudin2022}, as we summarize in Table~\ref{tbl:categorization} below: (1) \emph{Scope}, where explanations are provided either globally for the entire dataset, or locally for a single data record; (2) \emph{Method}, where explanations are produced either intrinsically as part of the model structure, or in a post-hoc manner based on the model outcomes; (3) \emph{Format}, where explanations are illustrated as a set of feature weights, a natural-language rationale, a series of examples, or visual displays; (4) \emph{Data}, where explanations are generated for explaining structured data or unstructured data.

\begin{table}[htbp]
\centering
\caption{Categorization of Explanations. Components Relevant to Our Paper are Marked in Bold.}
\label{tbl:categorization}
{\tabfont
\fitwidth{%
\begin{tabular}{>{\RaggedRight}p{0.225\linewidth} >{\RaggedRight}p{0.225\linewidth} >{\RaggedRight}p{0.225\linewidth} >{\RaggedRight\arraybackslash}p{0.18\linewidth}}
\toprule
\textbf{Scope} & \textbf{Method} & \textbf{Format} & \textbf{Data}\\
\midrule
\textbf{Global Explanation} (e.g., Importance Weight) &
\textbf{Intrinsic Explanation} (e.g., Attention Value) &
\textbf{Feature-Based} (e.g., Importance Weight) &
\textbf{Structured Data} (e.g., Tabular Data)\\
\addlinespace
\textbf{Local Explanation} (e.g., SHAP, LIME) &
\textbf{Post-hoc Explanation} (e.g., SHAP, LIME) &
\textbf{Natural Language} (e.g., Free-Text Rationale) &
Unstructured Data (e.g., Text, Image)\\
\addlinespace
& & \multicolumn{2}{l}{Example-Based (e.g., Prototype Learning)}\\
\addlinespace
& & \multicolumn{2}{l}{Visual Display (e.g., Grad-CAM)}\\
\bottomrule
\end{tabular}}}
\end{table}

\begin{figure}[htbp]
\centering
\begin{tikzpicture}[
  >=Latex, font=\scriptsize,
  data/.style={draw, rounded corners, inner sep=3pt},
  proc/.style={draw, rounded corners, fill=teal!15, thick, align=center, inner sep=4pt, font=\scriptsize\bfseries},
  outbox/.style={draw, rounded corners, inner sep=4pt, align=left, text width=5.6cm}
]
\node[data] (input) {%
  \setlength{\tabcolsep}{3pt}\renewcommand{\arraystretch}{1.0}%
  \begin{tabular}{ccccc}
    \textbf{C.Hist.} & \textbf{Inq.} & \textbf{Delq.} & \textbf{Util.} & \textbf{Score}\\
    \midrule
    5  & 1 & 0 & 0.26 & 691\\
    7  & 1 & 2 & 0.55 & 677\\
    13 & 4 & 7 & 0.83 & 625\\
  \end{tabular}%
};
\node[proc, right=6mm of input] (llm) {Rashomon\\LLM\,(EPR)};
\node[outbox, fill=blue!5, above right=1mm and 7mm of llm] (global) {%
  \textbf{Global:} Credit History 27\%, Inquiries 32\%, Delinquencies 20\%, Utilization 21\%.};
\node[outbox, fill=orange!12, below right=1mm and 7mm of llm] (local) {%
  \textbf{Local (record 1):} zero delinquencies and low utilization lift the predicted score; short credit history limits it.};
\draw[->, thick] (input) -- (llm);
\draw[->, thick] (llm.east) -- (global.west);
\draw[->, thick] (llm.east) -- (local.west);
\end{tikzpicture}
\caption{An Example of the RashomonLLM Workflow: from a tabular input, the Explanation--Prediction--Reflection loop produces both a global feature-contribution explanation and a local per-record rationale.}
\label{fig:workflow}
\end{figure}
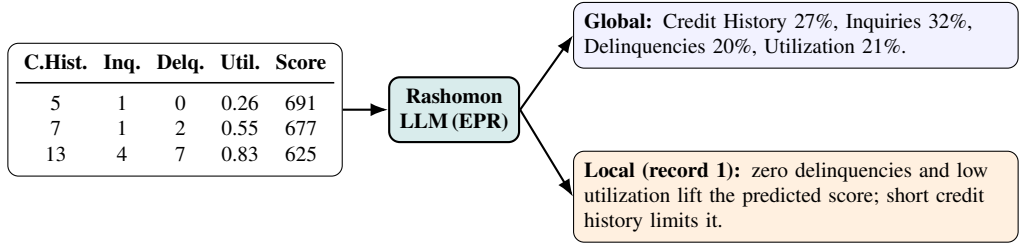

Among these categorizations, our framework applies at both the global and local scopes: it can summarize the feature relationships that govern an entire learning task (the importance weights in Figure~\ref{fig:workflow}), and it can equally generate a faithful rationale for an individual prediction (the per-exposure explanations illustrated in Table~\ref{tbl:kuaiexamples}). Our proposed method is also relevant to both intrinsic and post-hoc explanations, as it takes advantage of the benefits of both techniques in the sense that explanations are both part of the intrinsic learning process, and can be derived based on model outputs. Regarding format, we aim at producing feature-based explanations expressed as natural-language rationales, as our explanations take a semi-structured, natural-language form based on a template we provide a priori, capturing non-linear or conditional relationships while remaining directly readable. Finally, we focus only on structured, tabular data, while leaving the extension to unstructured data as future work. An example of our workflow is shown in Figure~\ref{fig:workflow}.

Besides the form of an explanation, we may also categorize it based on target: what an explanation claims about \citep{shmueli2010,lipton2018}, which includes four nested levels summarized in Table~\ref{tbl:levels}: (1) the fitted \emph{model} itself; (2) the \emph{observed data} on which the model was trained; (3) the underlying \emph{data population}, i.e., the distribution from which the data are drawn; and (4) the \emph{ground-truth relationship}, i.e., the true causal mechanism that generates the outcome. Each higher level demands a stronger warrant: an explanation of a model need not generalize to the population, and a population-level pattern need not coincide with the causal mechanism \citep{shmueli2010}.

\begin{table}[htbp]
\centering
\caption{Levels of Explanatory Target. The Levels Addressed by Our Paper are Marked in Bold.}
\label{tbl:levels}
{\tabfont\setstretch{1}
\fitwidth{%
\begin{tabular}{>{\RaggedRight}p{0.15\linewidth} >{\RaggedRight}p{0.225\linewidth} >{\RaggedRight}p{0.3\linewidth} >{\RaggedRight\arraybackslash}p{0.29\linewidth}}
\toprule
\textbf{Level} & \textbf{Explanatory Target} & \textbf{Warrant a Valid Claim Requires} & \textbf{Status in This Paper}\\
\midrule
\textbf{(1) Model} & the fitted predictor $f$ & faithful attribution to the model's decision & \textbf{In scope (primary):} each RashomonLLM run self-explains\\
\addlinespace
\textbf{(2) Observation} & the training sample $\{(x_i,y_i)\}$ & descriptive fidelity to the sample in hand & \textbf{In scope}\\
\addlinespace
\textbf{(3) Population} & the distribution $P(X,Y)$ & generalization beyond the sample (held-out fidelity) & \textbf{In scope (target):} via the Rashomon Explanation set\\
\addlinespace
(4) Causality & the causal mechanism & causal identification/intervention & Out of scope (not claimed)\\
\bottomrule
\end{tabular}}}
\end{table}

Each explanation produced by RashomonLLM is first of all a Level-1 statement about a single fitted model, and it is anchored to descriptive fidelity on the observed sample (Level~2). Our Rashomon Explanation set is the mechanism that helps us to achieve Level~3, since a feature that recurs across all near-optimal models in the Rashomon set is a property of the well-fitting model class on the population, not an artifact of any single fit \citep{fisher2019,semenova2022}. We explicitly do not claim Level~4 as we make no causal-identification or interventional claims. Therefore, the scope of this paper lies in its title: ``\emph{All Explanations are Wrong}'' restates the Level-4 point that no single explanation should be mistaken for the ground-truth mechanism \citep{box1976}; ``\emph{But Many Are Useful}'' is the Level-3 claim that explanations aggregated across the Rashomon set recover population-level structure faithful enough to be decision-relevant.

\section{Methodology}\label{sec:method}

\begin{figure}[htbp]
\centering
\includegraphics[width=\linewidth]{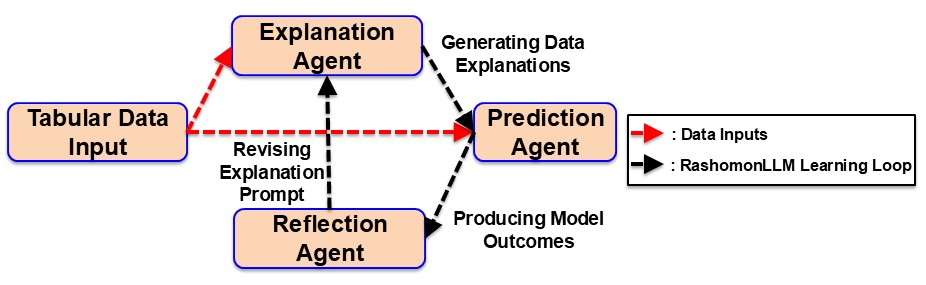}
\caption{Illustration of the Three Agents in Our Proposed RashomonLLM Method.}
\label{fig:agents}
\end{figure}

We now present the details of our proposed RashomonLLM method in this section to explore the entire Rashomon Explanation set. We adopt an LLM as the backbone model, since LLMs are capable of complex reasoning and deduction \citep[T.][]{brownT2020}. However, the challenges lie in hallucinations \citep{huangL2025} and context-sensitive decisions \citep[K.~E.][]{brownK2025}, which we discussed in Section~\ref{sec:rw-llm}. Our RashomonLLM method addresses these concerns based on the following two hypotheses: (1) Explanations, when aligned properly with prediction outcomes, can help improve the prediction performance and, in turn, the explanation quality; (2) an LLM agentic workflow that generates explanations across diverse configurations can explore the Rashomon Explanation set efficiently. These hypotheses motivate us to build an LLM agentic framework of ``Explanation--Prediction--Reflection'' (EPR), which consists of three LLM agents shown in Figure~\ref{fig:agents}. We first ground this design in organizational theory, then describe the three agents in detail, and we validate both hypotheses theoretically and empirically later in this paper.

\subsection{Organizational Motivations of Our Model Design}\label{sec:orgtheory}

Our agentic framework draws on three organizational theories with distinct, non-overlapping levels of analysis, so each yields one design principle of our proposed framework described in Table~\ref{tbl:theory}.

\textbf{Sensemaking Theory: What to Produce?} \citeauthor{weick1995}'s (\citeyear{weick1995}) theory holds that meaning-construction generates plausible, action-guiding accounts rather than a single objective truth, and is inherently plural; IS research accordingly argues that decision support should accommodate interpretive plurality rather than impose only one single solution \citep{sharma2003,volkoff2013}. Therefore, it is crucial to produce a set of explanations based on their capacity to guide prediction, precisely the ``usefulness'' criterion of our Rashomon Explanation in Definition~3.

\stat{Design Principle 1:} \emph{XAI systems should produce a set of action-guiding explanations rather than optimizing for a single explanation, reflecting the plural and pragmatic nature of sensemaking.}

\textbf{Double-Loop Learning Theory: How to Produce?} \citeauthor{argyris1997}'s (\citeyear{argyris1997}) theory distinguishes single-loop correction within a fixed framework from double-loop learning, which treats persistent errors as evidence that the framework is misspecified, so that it can be revised accordingly \citep{sein2011,kohli2004}. Therefore, our produced explanations need to be consistently updated in the learning process, so that we can minimize its misspecification error.

\stat{Design Principle 2:} \emph{XAI systems should incorporate a double-loop learning mechanism that continuously revises the explanation results in response to the observed prediction failures.}

\textbf{Complementarity Theory: How to Organize?} Complementarity theory \citep{milgrom1990} holds that two activities are complements when the marginal return to each rises with the level of the other, so a jointly designed system outperforms independent modules \citep{tanriverdi2005}. In our framework, since we hypothesize that the prediction task and the explanation task are mutually beneficial for each other, they should be optimized jointly rather than in isolation.

\stat{Design Principle 3:} \emph{XAI systems should jointly optimize explanation generation, outcome prediction, and framework reflection, rather than building independent modules.}

Putting them together, sensemaking sets the goal (a set of explanations), double-loop learning fixes the process (iterative revision), and complementarity determines the architecture (joint optimization). RashomonLLM, which is detailed in the remainder of this section, is one specific instantiation of these design principles that leads to various theoretical and empirical benefits.

\begin{table}[htbp]
\centering
\caption{Organizational Theory Motivations of Our Proposed RashomonLLM Method.}
\label{tbl:theory}
{\tabfont
\fitwidth{%
\begin{tabular}{>{\RaggedRight}p{0.42\linewidth} >{\RaggedRight}p{0.17\linewidth} >{\RaggedRight}p{0.19\linewidth} >{\RaggedRight\arraybackslash}p{0.2\linewidth}}
\toprule
\textbf{Theory} & \textbf{Level of Analysis} & \textbf{Design Principle} & \textbf{Design Components}\\
\midrule
Sensemaking \citep{weick1995} & Epistemological & Explanation Plurality & Rashomon Explanation\\
Double-Loop Learning \citep{argyris1997} & Process & Framework Revision & Reflection Agent\\
Complementarity \citep{milgrom1990} & Architecture & Joint Optimization & EPR Framework\\
\bottomrule
\end{tabular}}}
\end{table}

\subsection{Explanation Agent --- Generating Explanations via In-Context Learning}\label{sec:explanation}

The first component in our framework is the Explanation Agent, where we prompt the LLM to generate explanations based on the observed outcomes in the training data, similar to existing post-hoc explainers. However, the challenge is that we do not know the structure of the ground-truth explanations a priori, and consequently, we do not know what the best explanation prompt is. For example, if we prompt the LLM to focus only on the linear relationships, the explanations would overlook non-linear or conditional relationships between features; meanwhile, if we push the LLM to explore more complex relationships, it might not perform well on simple linear cases.

To tackle this challenge, rather than committing to a single prompt and hence a single explanation, we develop a new mechanism of \emph{LLM Feature Dropout}, where in each training iteration, we randomly drop 20\% of the features from the list of input features exposed to the LLM, forcing the agent to explain the outcome through different feature subsets across iterations. It serves two purposes: it prevents the agent from overfitting to a single dominant pattern, and it pushes the LLM to surface the interplay among feature combinations that would be omitted if the full feature set were always available. This mechanism is analogous to the feature subsampling used in random forests and other ensemble methods, while the difference is that we implement it by changing the prompting strategy rather than the data inputs. By varying the exposed features in this way, a single Explanation Agent eventually generates an element within the Rashomon Explanation set, according to Theorem~4 in Section~\ref{sec:theory-analysis}, and we therefore explore the Rashomon Explanation set with multiple, parallel Explanation Agents with different explanation prompts. These multiple initial templates are produced by LLM-enabled paraphrasing: the LLM rephrases a seed explanation prompt (detailed in Section~\ref{sec:prompts}) into semantically equivalent but lexically varied templates, each seeding a separate run of the EPR loop, so that the diversity of the Rashomon Explanation set arises from both the varied initial templates and feature dropout. Each explanation prompt is first initialized and then repeatedly refined by the Reflection Agent to better align with the underlying data distribution and the prediction task.

\subsection{Prediction Agent --- Predicting Model Outcomes Based on Generated Explanations}\label{sec:prediction}

Once we obtain the (initial) explanations produced by the Explanation Agent, we will next build the Prediction Agent to utilize these explanations for predicting the model outcomes. The prediction prompt is formatted in a manner that enables us to incorporate the explanation along with the training data to guide the reasoning process of LLMs in order to make accurate predictions. While there are many different ways to craft the prediction prompt that fit our framework well, the performance remains relatively steady across different prediction prompts in our empirical analysis. This is because the task of the Prediction Agent is relatively simple: it only needs to execute the instructions specified in the provided explanations to produce the prediction outcomes, and does not involve a complex reasoning process; as a result, we do not need any sophisticated prompting strategies, consistent with the finding that prompt sensitivity is low for simple tasks but high for complex, reasoning-intensive ones \citep{zhuo2024}.

\subsection{Reflection Agent --- Reflecting on Prediction Errors and Updating Explanation Format}\label{sec:reflection}

As we mentioned in Section~\ref{sec:explanation}, since we do not know the structure of ground-truth explanations a priori, it is crucial to implement an updating scheme to calibrate the prompt design and to push the LLM to get closer to the ground-truth structure and make correct predictions accordingly. To this end, we build the Reflection Agent, which aggregates all prediction errors made by the Prediction Agent and revises the explanation prompt accordingly. Specifically, we assume that the prediction mistakes made by the Prediction Agent are not because the prediction model is wrong (since the LLM is a powerful model to handle the prediction task), but because the explanations generated by the Explanation Agent do not match the actual data patterns, and as a result, provide wrong instructions to the Prediction Agent. We build the reflection prompt with the current explanation and a sample of the Prediction Agent's errors, so that the LLM would identify the responsible patterns and rewrite the explanation to correct them. The detailed template is shown in Section~\ref{sec:prompts}. Through repeated iterations among these three agents, we can simultaneously improve the explanation quality and the prediction performance over time, according to our experiment results and the theoretical foundation of Theorem~2, which we will further elaborate on later in this section.

\subsection{Summarization Module --- Aggregating the Rashomon Explanation Set}\label{sec:summarization}

Running the EPR loop across multiple initial templates and feature-dropout masks yields a collection of explanations. Downstream applications, meanwhile, require consolidating this set into a single coherent explanation. We perform this with a \emph{summarization module} that extracts the consensus, or majority interpretation, across the set. It proceeds in three steps, each completed by a specialized LLM agent. First, it decomposes each explanation into a collection of atomic claims of directional, conditional, or non-linear statements about features (e.g., ``high balance raises churn risk when tenure is low''), following the statement grammar of Section~\ref{sec:explanation}. Second, it tallies these claims by semantic equivalence, and matches each claim to a canonical proposition, recording its support as the fraction of explanations in the set whose claims include that proposition. Third, it retains propositions whose support exceeds a majority threshold $\kappa$ (we use standard majority $\kappa=0.5$) and discards those supported by only a small fraction, forming the final explanation.

This majority rule is consistent with our Rashomon analysis, as a claim that recurs across many independently seeded near-optimal explanations is far more likely to reflect a genuine relationship in the data than a template-specific one \citep{fisher2019}. It also explains the diminishing-returns structure documented in our experiments (Tables~\ref{tbl:robexpl}--\ref{tbl:robpred} and~\ref{tbl:kuaiksweep}): enlarging the set sharpens the estimated support of each claim, so the majority interpretation becomes more stable as members are added; once the set is large enough for the consensus to stabilize, additional members may no longer change the aggregated explanation, resulting in saturated prediction and explanation quality.

One might object that Theorem~1 also applies to this aggregated explanation, since it is itself a single explanation. We would like to clarify that first, the consensus is taken across independent, near-optimal explanations, so the error region on which any one of them is wrong shrinks as the set grows, and a larger distribution shift is required to expose an error; second, the usefulness guarantee of Definition~3 is an on-distribution property, whereas Theorem~1's failure mode is off-distribution. We additionally verify its robustness under distribution shift empirically through the prior-shift and time-based-split checks of Section~\ref{sec:kuairobustness}.

\subsection{Theoretical Analysis of Our Proposed Method}\label{sec:theory-analysis}

We now establish the theoretical properties of RashomonLLM. We denote by $\Phi(\cdot)$ the embedding map from explanations to $\mathbb{R}^{d}$. Our theoretical guarantees rest on the following assumptions.

\stat{Assumption~3 (Per-configuration stability).} For each run $c=(p,m)$ with initial template $p$ and feature-dropout mask $m$, we assume the expected one-step EPR update $T_{c}(e) := \mathbb{E}\!\left[\Phi(E') \mid \Phi(E)=e\right]$ over the LLM's stochastic decoding is a contraction, $\lVert T_{c}(e) - T_{c}(e') \rVert_{2} \leq k\lVert e - e' \rVert_{2}$ for some $k \in (0,1)$.

\stat{Assumption~4 (Expressiveness).} For any $E \in \mathcal{E}$ and any $\delta > 0$, there exist a prompt $p$ and temperature $\tau$ with $\Pr\!\left( \lVert \Phi(\text{LLM}(p,\tau)) - \Phi(E)\rVert_{2} \leq \delta \right) > 0$.

\stat{Assumption~5 (Coverage).} The family of fixed points $\{e^{*}_{c}\}_{c}$ is dense in the embedded Rashomon Explanation set: for every $E \in R_{explanation}(F,\theta)$ and every $\varepsilon>0$, there exist a configuration $c$ and an initialization under which the EPR process converges to $e^{*}_{c}$ with $\lVert e^{*}_{c} - \Phi(E)\rVert_{2} \leq \varepsilon$.

\stat{Assumption~6 (Injectivity).} For the embedding $\Phi$ defined on semantic equivalence classes, any distinct $E_{1},E_{2} \in R_{explanation}(F,\theta)$ satisfy $\lVert \Phi(E_{1}) - \Phi(E_{2})\rVert_{2} > \delta_{I} > 0$.

\stat{Theorem~4 (Characterization of RashomonLLM).} Under Assumptions~2--6, the RashomonLLM model optimized through the EPR loop with stopping criterion $L(f_{e}) \leq L(f^{*}) + \theta$ satisfies:

\emph{(i)~Soundness}: for any configuration $c$, if its fixed point $e^{*}_{c}$ meets the stopping criterion and the Rashomon radius satisfies $\theta \geq \psi^{-1}(\theta)$, then any $E^{*}$ with $\Phi(E^{*}) = e^{*}_{c}$ satisfies $E^{*} \in R_{explanation}(F,\theta)$.

\emph{(ii)~Convergence and Efficiency}: for each configuration $c$, the deterministic iterates of the mean update, $\bar{e}^{c}_{t+1} = T_{c}(\bar{e}^{c}_{t})$ started at $\bar{e}^{c}_{0} = e^{c}_{0}$, converge geometrically to $e^{*}_{c}$, with $\lVert \bar{e}^{c}_{t} - e^{*}_{c}\rVert_{2} \leq k^{t}\lVert e^{c}_{0} - e^{*}_{c}\rVert_{2}$, reaching an $\varepsilon$-approximation within $\lceil \log(\lVert e^{c}_{0}-e^{*}_{c}\rVert_{2}/\varepsilon)/\log(1/k)\rceil$ iterations.

\emph{(iii)~Completeness}: for every $E \in R_{explanation}(F,\theta)$ and every $\varepsilon>0$, some configuration $c$ yields an output $E'$ with $\Phi(E')=e^{*}_{c}$ and $\lVert \Phi(E') - \Phi(E)\rVert_{2} \leq \varepsilon$.

\stat{Proof.} Fix a configuration $c$. By Assumption~3, $T_{c}$ is a contraction, so the Banach fixed-point theorem gives a unique fixed point $e^{*}_{c}$ and the geometric rate of the mean iterates in (ii); the iteration count follows from solving $k^{t}\lVert e^{c}_{0}-e^{*}_{c}\rVert_{2} \leq \varepsilon$. The $O(\sigma/(1-k))$ concentration of the realized iterates is the standard bound for a contraction driven by zero-mean, variance-$\sigma^{2}$ perturbations. For (i), the stopping criterion places $f_{e^{*}_{c}}$ in $R_{set}(F,\theta)$, and by the fidelity coupling of Assumption~2 the associated explanation satisfies $d(E^{*},E_{S}) \leq \psi^{-1}(\theta) \leq \theta$ (the last step using $\theta \geq \psi^{-1}(\theta)$), so $E^{*} \in R_{explanation}(F,\theta)$. For (iii), fix $E \in R_{explanation}(F,\theta)$ and $\varepsilon>0$; by Assumption~5 some configuration $c$ has $\lVert e^{*}_{c}-\Phi(E)\rVert_{2} \leq \varepsilon$, while Assumption~6 keeps distinct members $\delta_{I}$-separated so the covering is non-degenerate. Finally, since the realized iterates concentrate within $O(\sigma/(1-k))$ of $e^{*}_{c}$ as established above, the soundness (i) and coverage (iii) conclusions transfer to the realized outputs up to this decode-noise radius, which our stability diagnostic (Section~\ref{sec:kuailive}) confirms is small relative to $\theta$. \hfill Q.E.D.

\stat{Theorem~4 In Plain Language:} each run of RashomonLLM (each template/dropout) converges quickly to its own explanation, and by varying the run configuration the method recovers every explanation in the Rashomon Explanation set: convergence is per run, coverage is across runs.

These assumptions are explicit and verifiable. \emph{Stability} (Assumption~3) is diagnosed directly by the per-iteration displacement $\lVert e_{t+1}-e_{t}\rVert_{2}$, whose geometric decay we report for our industrial study (Section~\ref{sec:kuailive}, Table~\ref{tbl:kuaicontract}). \emph{Expressiveness} (Assumption~4) requires only that some prompt--temperature setting can get arbitrarily close to a target explanation, which is easily met by a modern LLM sampled at nonzero temperature over the diverse, paraphrase-generated initial templates of Section~\ref{sec:explanation}. \emph{Coverage} (Assumption~5) is a stronger condition on the EPR dynamics, and we assess it empirically through the diversity and spread of the recovered explanations. \emph{Injectivity} (Assumption~6) holds for any non-degenerate embedding that maps semantically distinct explanations to distinct points, and is verified by checking that the pairwise embedding distances among the recovered explanations stay bounded away from zero.

\subsection{Batch LLM Learning --- An Extension of Our Proposed Method to Enhance Scalability}\label{sec:batch}

While our proposed RashomonLLM method is effective in tackling the problems in existing XAI methods, both from the theoretical perspective (as we have illustrated in Section~\ref{sec:theory-analysis}) and from the empirical perspective (as we will demonstrate in Section~\ref{sec:experiments} next), an important practical challenge still exists for our proposed method: the limit of the context window. State-of-the-art LLMs, such as GPT or Claude, can handle context windows on the order of a million tokens, while the sizes of data that RashomonLLM needs to handle are typically at the scale of millions or even billions for industrial platforms. As a result, we cannot feed the entire dataset into the LLM in a trivial manner. In addition, previous literature has shown that when the context history gets longer, LLMs are more likely to suffer from long-context degradation \citep{liu2024lost}, where the model attends less reliably to information buried deep in a long context.

\begin{figure}[htbp]
\centering
\resizebox{\linewidth}{!}{%
\begin{tikzpicture}[
  >=Latex, font=\footnotesize,
  box/.style={draw, rounded corners, align=center, inner sep=3pt, minimum height=7mm, minimum width=15mm},
  llm/.style={box, fill=teal!15, thick},
  mon/.style={box, draw=red, text=red, thick},
  bt/.style={box, fill=orange!12}
]
\node[llm] (llm) {Backbone\\LLM};
\node[box, right=9mm of llm] (p1) {Data\\Prompting};
\node[box, right=9mm of p1] (p2) {Data\\Prompting};
\node[right=5mm of p2] (d) {$\cdots$};
\node[box, right=5mm of d] (pN) {Data\\Prompting};
\node[llm, right=9mm of pN] (out) {Data-Enhanced\\LLM};
\node[bt, below=8mm of p1] (b1) {Data Batch 1};
\node[bt, below=8mm of p2] (b2) {Data Batch 2};
\node[bt, below=8mm of pN] (bN) {Data Batch N};
\node[llm, left=9mm of b1] (input) {Input\\Data};
\node[mon, above=7mm of p2] (mon) {Data Monitor};
\draw[->,thick] (llm)--(p1);
\draw[->,thick] (p1)--(p2);
\draw[->,thick] (p2)--(d);
\draw[->,thick] (d)--(pN);
\draw[->,thick] (pN)--(out);
\draw[->,thick] (b1)--(p1);
\draw[->,thick] (b2)--(p2);
\draw[->,thick] (bN)--(pN);
\path (input.south) ++(0,-4mm) coordinate (bus);
\draw[thick] (input.south) -- (bus);
\draw[thick] (bus) -- (bus -| bN.south);
\draw[->,thick] (bus -| b1.south) -- (b1.south);
\draw[->,thick] (bus -| b2.south) -- (b2.south);
\draw[->,thick] (bus -| bN.south) -- (bN.south);
\draw[->,red,thick] (mon)--(p1);
\draw[->,red,thick] (mon)--(p2);
\draw[->,red,thick] (mon)--(pN);
\end{tikzpicture}%
}
\caption{Illustration of Our Batch LLM Learning Mechanism for Handling Large Input Sizes.}
\label{fig:batch}
\end{figure}
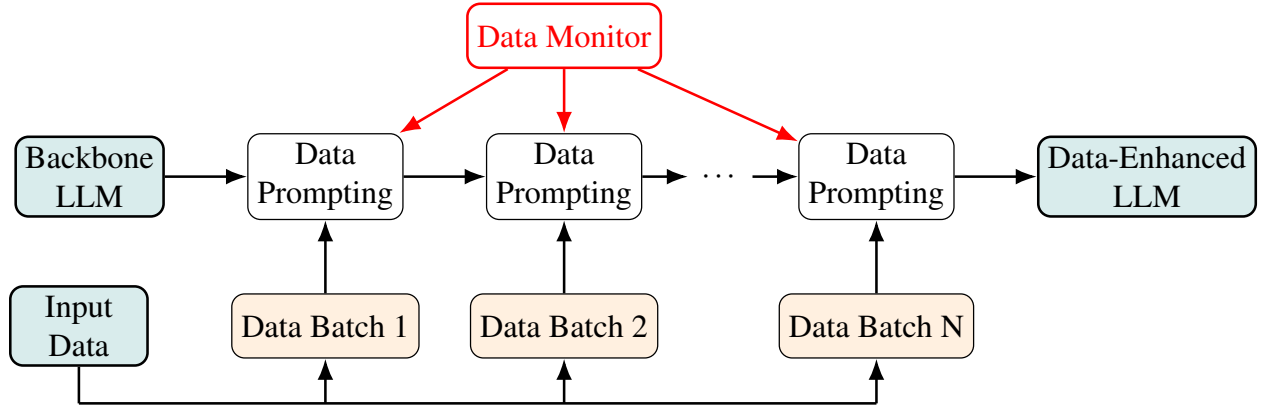

To address these challenges, we develop a new \emph{Batch LLM Learning} mechanism, where we segment the input training dataset into multiple ``batches'' and feed them into the LLM sequentially, one batch at a time, so that each batch incrementally updates a running summary of the data that is carried forward in the prompt. A practical concern with in-context batching is that, as the running context grows, the LLM may lose track of information from earlier batches. As a lightweight safeguard, we introduce an ``\emph{LLM Data Monitor}'': we carry a small set of sentinel rows sampled from earlier batches forward in the prompt, and after feeding each batch we ask the LLM to reproduce these sentinel rows together with rows from the current batch, re-prompting whenever the reproduced values diverge from the originals, with a final check after the last batch. The entire workflow of our Batch LLM Learning mechanism is explained in Figure~\ref{fig:batch}, where by prompting the backbone LLM through a series of data batches we obtain a ``\emph{Data-Enhanced LLM},'' i.e., the frozen LLM backbone conditioned on a running summary refined over all data batches. This conditioned LLM then becomes the foundation for the agents in our RashomonLLM method.

Another unique design in our method is a mini-batch processing scheme, loosely analogous to the Batch Normalization \citep{ioffe2015} technique. Specifically, we apply the Reflection Agent to each batch of the data to obtain the outputs on that batch accordingly, rather than applying it directly to the entire dataset. As a result, we will first reflect on the errors made within each data batch to make recommendations to revise our explanation prompt, and then aggregate all these recommended revisions together to obtain the final updated explanation prompt. While the batch size is clearly an important hyperparameter to be determined for our RashomonLLM model, in our experiments discussed in the next section, we select it as a number that is slightly below the LLM context window (i.e., 100,000 tokens for GPT-4o with a context window of 128,000 tokens), so that we can achieve a good balance between the model performance and the computational efficiency.

\subsection{Agent Prompt Templates}\label{sec:prompts}

For completeness, Table~\ref{tbl:prompts} records the prompt templates for the three agents across our three applications, namely Bank Churn classification, HCT Survival regression, and the KuaiLive CTR prediction in Section~\ref{sec:kuailive}. We use angle brackets ($\langle\cdot\rangle$) to denote runtime substitutions (serialized features, generated explanations, or labels). We also evaluated alternative phrasings of the Prediction Agent prompt (varying the wording and the order of the explanation and data blocks) and found predictions stable across them, consistent with its role of executing the supplied patterns rather than reasoning open-endedly (Section~\ref{sec:theory-analysis}). In the synthetic credit-score study (Section~\ref{sec:useful}), the five strategies share the Prediction Agent prompt and differ only in the explanation supplied: no explanation; the ground-truth explanation perturbed with 10\%, 5\%, and 1\% feature-attribution noise; and the exact ground-truth, with the noise injected by randomly reweighting feature importance.

\begin{table}[!ht]
\centering
\caption{Prompt Templates for the Three RashomonLLM Agents (Explanation, Prediction, Reflection) Across the Three Applications. Angle brackets ($\langle\cdot\rangle$) denote runtime substitutions.}
\label{tbl:prompts}
{\tabfont\setstretch{1}
\fitwidth{%
\begin{tabular}{l >{\RaggedRight\arraybackslash}p{0.80\linewidth}}
\toprule
\textbf{Agent} & \textbf{Prompt Template}\\
\midrule
\multicolumn{2}{l}{\textbf{Bank Churn} --- binary classification, target \textit{Exited}}\\
\textbf{Explanation} & ``Examine the following labeled examples and describe, as a short numbered list of conditional rules, the patterns that determine $\langle$target$\rangle$. Cite specific features and their value ranges. $\langle$data\_record$\rangle$''\\
\textbf{Prediction} & ``Based on the following explanations: $\langle$generated\_explanations$\rangle$, predict if each customer in the following numbered list will exit (1 for Yes, 0 for No). Your analysis must be strictly based on the provided patterns. $\langle$data\_record$\rangle$''\\
\textbf{Reflection} & ``Here are instances the current explanation misclassified, with their true labels: $\langle$error\_cases$\rangle$. Identify which rules are responsible and rewrite the explanation to correct these errors without discarding rules that already work.''\\
\midrule
\multicolumn{2}{l}{\textbf{HCT Survival} --- regression, target \textit{efs\_time} (months)}\\
\textbf{Explanation} & ``Examine the following labeled examples and describe, as a short numbered list of conditional rules, the patterns that determine $\langle$target$\rangle$, the event-free survival time in months. Cite specific features and their value ranges. $\langle$data\_record$\rangle$''\\
\textbf{Prediction} & ``Based on the following explanations: $\langle$generated\_explanations$\rangle$, predict the event-free survival time (in months) for each patient in the following numbered list. Your analysis must be strictly based on the provided patterns. $\langle$data\_record$\rangle$''\\
\textbf{Reflection} & ``Here are instances the current explanation predicted with the largest error, with their true survival times: $\langle$error\_cases$\rangle$. Identify which rules are responsible and rewrite the explanation to correct these errors without discarding rules that already work.''\\
\midrule
\multicolumn{2}{l}{\textbf{KuaiLive} --- industrial CTR, target \textit{clicked}}\\
\textbf{Explanation} & ``Examine the following labeled exposure examples and describe, as a structured set of conditional rules, the patterns that distinguish a click from no click---citing user engagement history, streamer content niche and popularity, and time-of-day. $\langle$data\_record$\rangle$''\\
\textbf{Prediction} & ``Based on the following explanations: $\langle$generated\_explanations$\rangle$, predict for each exposure in the following numbered list whether the user clicks (1 for click, 0 for no click). Your analysis must be strictly based on the provided patterns. $\langle$data\_record$\rangle$''\\
\textbf{Reflection} & ``Here are exposures the current explanation mispredicted, with their true labels: $\langle$error\_cases$\rangle$. Identify which rules are responsible and rewrite the explanation to correct these errors without discarding rules that already work.''\\
\bottomrule
\end{tabular}}}
\end{table}

\subsection{Technical Novelty of RashomonLLM}\label{sec:novelty}

To conclude, we summarize the key technical novelties of RashomonLLM over existing methods.

\begin{itemize}[leftmargin=2em]
\item \textbf{A collection of explanations instead of one.} Prior XAI methods commit to a single explanation, which is fragile under distribution shift (Theorem~1). RashomonLLM instead constructs and aggregates a set of near-optimal, faithful explanations (Rashomon Explanation).
\item \textbf{Natural-language explanation form.} Existing attribution methods return scalar feature weights (e.g., SHAP, LIME) that cannot represent conditional or non-linear structure, while RashomonLLM produces semi-structured natural-language rationales that express directional, conditional, and non-linear structures while remaining directly readable by stakeholders.
\item \textbf{Explanation coupled with prediction.} Post-hoc explainers are decoupled from the prediction model and cannot improve it. Through the EPR loop, RashomonLLM couples the two, so that producing a faithful explanation lowers prediction error (Theorem~2), therefore resolving the accuracy--explainability trade-off.
\item \textbf{LLM Feature Dropout.} This new technique diversifies the explanation set by perturbing the subset of features exposed to the model, enabling us to realize an ensemble effect.
\item \textbf{Batch LLM Learning.} This new mechanism scales the workflow to data far beyond the LLM context window through in-context batching.
\end{itemize}

To our knowledge, RashomonLLM is the first method to make explanation and prediction mutually reinforcing while exploring the Rashomon Explanation set, with per-run convergence backed by formal guarantees (Theorems~1--4) and full coverage of the set under the conditions we assess empirically. We now demonstrate its empirical benefits.

\section{Offline Experiments}\label{sec:experiments}

\subsection{Datasets}\label{sec:datasets}

To demonstrate the effectiveness of our proposed RashomonLLM method, we implement it based on GPT-4o on two benchmark datasets associated with recent competitions on Kaggle:

\begin{itemize}[leftmargin=2em]
\item \textbf{Bank Churn Dataset},\footnote{Binary Classification with a Bank Churn Dataset (Kaggle Playground Series, Season 4, Episode 1): \url{https://www.kaggle.com/competitions/playground-series-s4e1}.} which is a binary classification task for predicting whether a customer will churn in banking. The dataset contains the customer demographics (e.g., Gender, Age), relationship with the bank (e.g., Number of Products), and financial status (e.g., Salary).
\item \textbf{HCT Survival Dataset},\footnote{CIBMTR --- Equity in Post-HCT Survival Predictions (Kaggle): \url{https://www.kaggle.com/competitions/equity-post-HCT-survival-predictions}.} which is a regression task for predicting the months of transplant survival for allo-HCT patients. The dataset contains information about Patient Characteristics (e.g., Demographics, Health Status) and Transplant Specifics (e.g., Matching, Source).
\end{itemize}

We select these two datasets for the following reasons. First, they represent two distinct machine learning tasks (regression and classification) with significantly different business applications and data distributions; as a result, the performance on both datasets can demonstrate the generalizability and robustness of our proposed method. Second, these two datasets were recently released on Kaggle as machine learning competitions in which many data science teams participated and submitted strong solutions; by comparing with the winning solutions for both tasks, we are able to demonstrate the superior performance and the power of our proposed model. Third, the GPT-4o version we use (\texttt{gpt-4o-2024-05-13}) has a knowledge cutoff of October 2023, which precedes the public release of both datasets; we deliberately select such post-cutoff datasets so that the backbone LLM cannot have memorized the data or the winning competition solutions during pretraining, ensuring a clean, contamination-free evaluation. We include details about the size and specific features of both datasets in Table~\ref{tbl:datasets} below, and we report the official competition metrics (AUC for Bank Churn and the stratified C-index for HCT) on the Kaggle leaderboard test split, so they are directly comparable to the winning competition solutions; because the competition test labels are withheld, all remaining label-dependent metrics, including Accuracy, F1-Score, RMSE, MAE, and the faithfulness measures, are computed on an 80/20 split of the released training data, with the winning solutions re-run on the same split, and part of that split also serving as the validation set for the EPR stopping rule. For Bank Churn, we exclude the identifier columns (\emph{id}, \emph{CustomerId}, and \emph{Surname}) from all models, dropping \emph{Surname} in particular to avoid leakage-adjacent signal.

\begin{table}[htbp]
\centering
\caption{Detailed Information about the Datasets Used in Our Experiments; Target Variable in Bold.}
\label{tbl:datasets}
{\tabfont\setstretch{1}
\fitwidth{%
\begin{tabular}{>{\RaggedRight}p{0.15\linewidth} >{\RaggedRight}p{0.10\linewidth} >{\RaggedRight\arraybackslash}p{0.66\linewidth}}
\toprule
\textbf{Dataset} & \textbf{Data Size} & \textbf{Feature Information}\\
\midrule
Bank Churn\newline \footnotesize(\href{https://www.kaggle.com/competitions/playground-series-s4e1/data}{Kaggle Playground S4E1})
& 165,034
& \textit{Customer ID}: A unique identifier for each customer \newline
\textit{Surname}: the customer's last name \newline
\textit{Credit Score}: a numerical value representing the customer's credit score \newline
\textit{Geography}: the country where the customer resides \newline
\textit{Gender}: the customer's gender (Male or Female) \newline
\textit{Age}: the customer's age \newline
\textit{Tenure}: the number of years the customer has been with the bank \newline
\textit{Balance}: the customer's account balance \newline
\textit{NumOfProducts}: the number of bank products the customer uses \newline
\textit{HasCrCard}: whether the customer has a credit card (1 = yes, 0 = no) \newline
\textit{IsActiveMember}: whether the customer is active (1 = yes, 0 = no) \newline
\textit{EstimatedSalary}: the estimated salary of the customer \newline
\textbf{\emph{Exited}: whether the customer has churned (1 = yes, 0 = no)}.\\
\addlinespace
HCT Survival\newline \footnotesize(\href{https://www.kaggle.com/competitions/equity-post-HCT-survival-predictions/data}{Kaggle Equity Post-HCT})
& 28,800
& \textit{Primary Disease and Risk Scores}: 6 variables \newline
\textit{Comorbidity Indicators}: 13 categorical variables \newline
\textit{HLA Matching}: 17 variables (high- and low-resolution matches across HLA loci) \newline
\textit{T-cell Epitope (TCE) Matching}: 3 variables \newline
\textit{Transplant and Conditioning Specifics}: 12 variables \newline
\textit{Donor and Recipient Characteristics}: 6 variables \newline
\textit{efs}: the event indicator (1 = event observed, 0 = right-censored), used together with the target to compute the stratified C-index \newline
\textbf{\emph{efs\_time}: time to event-free survival (in months)}.\\
\bottomrule
\end{tabular}}}
\end{table}

\subsection{Baseline Models and Evaluation Metrics}\label{sec:baselines}

We compare our RashomonLLM method with the following four groups of baselines:

\begin{itemize}[leftmargin=2em]
\item \textbf{State-of-the-art XAI Models}, including \emph{Sparse Decision Tree} \citep{xin2022} and \emph{Sparse AutoEncoder} \citep{huben2023}. The former learns a decision tree model using only a small subset of features with concise rules. The latter learns a sparse, overcomplete dictionary over the black-box model's internal activations and attributes importance to the input features most strongly associated with the active dictionary units, adapting this language-model interpretability technique to the tabular setting.
\item \textbf{LLM-based Post-hoc Explainer}, which is a single-shot LLM (GPT-4o) prompted to produce a feature-importance explanation for the task directly. This baseline represents the prevailing way LLMs are currently used for explanation \citep{kroeger2023,ajwani2024}.
\item \textbf{Black-box Deep Learning Models for Tabular Data}, including \emph{TabPFN-3} \citep{tabpfn3_2026}, the latest release of the TabPFN tabular foundation model \citep{hollmann2025}, and \emph{TabTransformer} \citep{huangX2020}. The former is pre-trained on diverse synthetic datasets with a Transformer architecture for performing instant inference on new tabular data. The latter leverages a self-attentive Transformer architecture to transform categorical features into contextual embeddings for capturing complex relationships within the tabular data.
\item \textbf{Winning solutions on Kaggle}, the first-prize model for each task; their official AUC/C-index are the leaderboard scores, and their remaining metrics are re-computed on the same split as our models, with Accuracy and F1-Score obtained at a 0.5 threshold.
\end{itemize}

We select two dimensions of evaluation metrics, focusing on explanation quality and prediction accuracy. Because our notion of faithfulness is closeness to the true feature--outcome relationships (Definition~3), which are unobserved on real data, we assess it alternatively in two ways. First, where the ground truth is known, e.g., on the synthetic data of Section~\ref{sec:wrong}, we measure faithfulness directly as the rank correlation between an explanation and the true feature importances. Second, on the real-world datasets with no ground-truth, we use the following established, model-agnostic faithfulness metrics from the XAI literature \citep{nauta2023} that probe whether an explanation's identified drivers genuinely determine the observed outcome: \emph{Single Deletion}, the drop in task performance when a single highly ranked feature is removed, which is large only if that feature truly drives the outcome; and \emph{Randomization Check}, which verifies that the explanation does not assign importance to features that are in fact non-influential, so that perturbing them leaves predictions essentially unchanged. These perturbation-based metrics are well-established and empirically benchmarked measures of explanation faithfulness \citep{hooker2019}, and they align directly with faithfulness in our Definition~3. In all tables, both metrics are reported as the change in the reference predictor's performance relative to its unperturbed baseline, i.e., the change in AUC for the classification task and the change in normalized RMSE for the regression task.

The second group of metrics for prediction accuracy includes RMSE (Root Mean Square Error), MAE (Mean Absolute Error), Accuracy, and F1-Score, which are standard metrics for measuring the performance of prediction tasks and classification tasks, respectively. Additionally, we include the evaluation metrics of AUC and C-index used in both Kaggle competitions. For the HCT Survival task, event-free survival time (\emph{efs\_time}) is a right-censored time-to-event outcome; we regress on it directly and report RMSE/MAE, treating censored times as observed, following the standard practice of casting survival prediction as regression on the observed time \citep{wang2017survival}. Statistical significance is assessed against each baseline with a Holm--Bonferroni correction for multiple comparisons across baselines, using paired $t$-tests for the instance-decomposable metrics (Accuracy, MSE/RMSE, and MAE), the bootstrap for F1, the race-group--stratified C-index, and the faithfulness metrics (Single Deletion and Randomization Check), and DeLong's test for AUC. Because the competition test labels are withheld, all tests are computed on our labeled 80/20 split. The results will be reported in the next section.

\subsection{Main Results}\label{sec:mainresults}

We present the main experiment results and empirical findings in this section. As we show below in Table~\ref{tbl:explquality}, RashomonLLM generates explanations with significantly better quality compared to the XAI baselines (Sparse Decision Tree, Sparse AutoEncoder) and to the single-shot LLM post-hoc explainer, confirming that the gain stems from our EPR learning loop, rather than from prompting an LLM for explanations per se. Additionally, on the synthetic data in Section~\ref{sec:wrong}, RashomonLLM attains a Spearman correlation of 0.97 with the ground-truth importances, versus 0.71 for SHAP on the black-box models (both computed as the mean per-instance rank correlation over the $n=4$ generative features), confirming that our explanations track the true data-generating patterns and validating the proxy metrics used on the real-world datasets.

In addition, as illustrated in Table~\ref{tbl:predperf}, RashomonLLM also significantly improves the prediction and classification performance, when utilizing these generated explanations as part of model inputs, outperforming all state-of-the-art baselines, including the best Kaggle solutions. These performance improvements are consistent and significant across both datasets, all evaluation metrics, and baseline models, illustrating the robustness of our findings. These results not only demonstrate the superiority and validity of RashomonLLM, but also reaffirm the value of XAI in prediction and classification tasks, since they reflect a complementarity between explanation and prediction, where equipping the model to explain itself improves its accuracy; meanwhile, by aligning with the prediction target, we can also obtain more ``useful'' explanations in the Rashomon Explanation set, which will be beneficial for the downstream business applications.

\begin{table}[htbp]
\centering
\caption{Comparison of the Explanation Quality versus XAI Baselines. ***p$<$0.01.}
\label{tbl:explquality}
{\tabfont
\fitwidth{%
\begin{tabular}{lcccc}
\toprule
\textbf{Datasets} & \multicolumn{2}{c}{\textbf{Bank Churn (AUC)}} & \multicolumn{2}{c}{\textbf{HCT Survival (RMSE)}}\\
\cmidrule(lr){2-3}\cmidrule(lr){4-5}
\textbf{Evaluation Metrics} & \textbf{Randomization Check ($\downarrow$)} & \textbf{Single Deletion ($\uparrow$)} & \textbf{Randomization Check ($\downarrow$)} & \textbf{Single Deletion ($\uparrow$)}\\
\midrule
\textbf{RashomonLLM} & \textbf{0.0245***} & \textbf{0.0386***} & \textbf{0.0417***} & \textbf{0.0341***}\\
Sparse Decision Tree & 0.0251 & 0.0291 & NA & NA\\
Sparse AutoEncoder & 0.0259 & 0.0294 & 0.0510 & 0.0250\\
LLM Post-hoc Explainer & 0.0308 & 0.0229 & 0.0521 & 0.0279\\
\bottomrule
\end{tabular}}}

{\footnotesize\raggedright \textit{Note.} The Sparse Decision Tree method only handles binary labels and cannot work for the HCT Survival Dataset.\par}
\end{table}

\begin{table}[htbp]
\centering
\caption{Comparison of the Prediction Performance versus Baseline Models. ***p$<$0.01.}
\label{tbl:predperf}
{\tabfont\setlength{\tabcolsep}{4pt}
\fitwidth{%
\begin{tabular}{lcccccc}
\toprule
\textbf{Datasets} & \multicolumn{3}{c}{\textbf{Bank Churn}} & \multicolumn{3}{c}{\textbf{HCT Survival}}\\
\cmidrule(lr){2-4}\cmidrule(lr){5-7}
\textbf{Evaluation Metrics} & \textbf{Accuracy} & \textbf{F1-Score} & \textbf{AUC} & \textbf{RMSE} & \textbf{MAE} & \textbf{C-index}\\
\midrule
\textbf{RashomonLLM} & \textbf{0.921***} & \textbf{0.698***} & \textbf{0.912***} & \textbf{19.848***} & \textbf{15.502***} & \textbf{0.709***}\\
Sparse Decision Tree & 0.852 & 0.569 & 0.892 & NA & NA & NA\\
Sparse AutoEncoder & 0.864 & 0.580 & 0.901 & 22.530 & 16.886 & 0.694 \\
TabPFN-3 & 0.835 & 0.582 & 0.901 & 21.045 & 17.009 & 0.693 \\
TabTransformer & 0.825 & 0.484 & 0.898 & 27.702 & 21.493 & 0.691 \\
Best Solution on Kaggle & 0.902 & 0.653 & 0.906 & 20.372 & 15.807 & 0.701 \\
\bottomrule
\end{tabular}}}

{\footnotesize\raggedright \textit{Note.} The Sparse Decision Tree method only handles binary labels and cannot work for the HCT Survival Dataset.\par}
\end{table}

\subsection{Robustness Check and Additional Analysis}\label{sec:robustness}

Besides those main results, we have also conducted ablation studies and additional robustness experiments in this section, which we summarize as follows:

\begin{itemize}[leftmargin=2em]
\item \textbf{Ablation of RashomonLLM Components.} To isolate the contribution of each design element, we compare the full model against three ablated variants. \emph{(i)~Predict-Only}, where the LLM predicts directly from raw features, isolating the value that self-explanation brings to prediction. \emph{(ii)~Without Reflection}, where we retain only the Explanation and Prediction Agents in the iterative loop with fixed prompts, isolating the value of error-driven prompt revision. \emph{(iii)~Without Iteration}, where we run the three agents in a single pass without updating, isolating the value of the EPR loop. As reported in Table~\ref{tbl:ablpred} and Table~\ref{tbl:ablexpl}, all these ablations significantly degrade prediction performance, and the two explanation-producing variants further degrade explanation quality, relative to the full model, confirming that each component contributes to RashomonLLM's effectiveness.
\item \textbf{Number of Explanation Templates.} We vary the number of explanation templates that we use (i.e., the number of explanations that we will explore in the Rashomon Explanation set) to see how it will change the model performance. As shown in Table~\ref{tbl:robexpl} and Table~\ref{tbl:robpred}, increasing the number of explanations will be helpful for both the prediction and the explanation performance; however, when the explanation number reaches a certain threshold (i.e., 50 for both datasets), the performance improvements will only be marginal, indicating that this is a ``sweet spot'' for balancing between the model performance and computational efficiency.
\item \textbf{Temperature.} The temperature (ranging from 0 to 1) controls the randomness and creativity of the model's output. As we show in Table~\ref{tbl:robexpl} and Table~\ref{tbl:robpred}, our RashomonLLM model remains relatively steady across different selections of the temperature values.
\item \textbf{Batch Size.} We examine how different values of the batch size will potentially change the model performance. As we show in Table~\ref{tbl:robexpl} and Table~\ref{tbl:robpred}, RashomonLLM remains relatively steady across different selections of the batch sizes, suggesting that, since performance is stable across batch sizes, we may select the largest batch that fits the LLM context window to maximize training efficiency without sacrificing model performance.
\end{itemize}

\begin{table}[htbp]
\centering
\caption{Ablation Study: Prediction Performance of RashomonLLM Versus Its Variants. ***p$<$0.01.}
\label{tbl:ablpred}
{\tabfont\setlength{\tabcolsep}{4pt}
\fitwidth{%
\begin{tabular}{lcccc}
\toprule
\textbf{Variant} & \multicolumn{2}{c}{\textbf{Bank Churn}} & \multicolumn{2}{c}{\textbf{HCT Survival}}\\
\cmidrule(lr){2-3}\cmidrule(lr){4-5}
 & \textbf{Accuracy} & \textbf{F1-Score} & \textbf{RMSE} & \textbf{MAE}\\
\midrule
\textbf{RashomonLLM (full)} & \textbf{0.921***} & \textbf{0.698***} & \textbf{19.848***} & \textbf{15.502***}\\
Predict-Only & 0.903 & 0.651 & 21.339 & 16.238\\
Without Reflection & 0.838 & 0.601 & 26.369 & 21.664\\
Without Iteration & 0.736 & 0.433 & 38.372 & 25.893\\
\bottomrule
\end{tabular}}}
\end{table}

\begin{table}[htbp]
\centering
\caption{Ablation Study: Explanation Quality of RashomonLLM Versus Its Variants. ***p$<$0.01.}
\label{tbl:ablexpl}
{\tabfont\setlength{\tabcolsep}{4pt}
\fitwidth{%
\begin{tabular}{lcccc}
\toprule
\textbf{Variant} & \multicolumn{2}{c}{\textbf{Bank Churn (AUC)}} & \multicolumn{2}{c}{\textbf{HCT Survival (RMSE)}}\\
\cmidrule(lr){2-3}\cmidrule(lr){4-5}
 & \textbf{Randomization Check ($\downarrow$)} & \textbf{Single Deletion ($\uparrow$)} & \textbf{Randomization Check ($\downarrow$)} & \textbf{Single Deletion ($\uparrow$)}\\
\midrule
\textbf{RashomonLLM (full)} & \textbf{0.0245***} & \textbf{0.0386***} & \textbf{0.0417***} & \textbf{0.0341***}\\
Without Reflection & 0.0306 & 0.0101 & 0.0498 & 0.0268\\
Without Iteration & 0.0538 & 0.0319 & 0.0636 & 0.0317\\
\bottomrule
\end{tabular}}}
\end{table}

\begin{table}[htbp]
\centering
\caption{Comparison of the Explanation Quality for Robustness Check.}
\label{tbl:robexpl}
{\tabfont\setlength{\tabcolsep}{4pt}
\fitwidth{%
\begin{tabular}{lcccc}
\toprule
\textbf{Datasets} & \multicolumn{2}{c}{\textbf{Bank Churn (AUC)}} & \multicolumn{2}{c}{\textbf{HCT Survival (RMSE)}}\\
\cmidrule(lr){2-3}\cmidrule(lr){4-5}
\textbf{Evaluation Metrics} & \textbf{Randomization Check ($\downarrow$)} & \textbf{Single Deletion ($\uparrow$)} & \textbf{Randomization Check ($\downarrow$)} & \textbf{Single Deletion ($\uparrow$)}\\
\midrule
Number of Explanations = 10 & 0.0259 & 0.0368 & 0.0449 & 0.0328\\
Number of Explanations = 20 & 0.0251 & 0.0381 & 0.0422 & 0.0339\\
Number of Explanations = 50 (default) & 0.0245 & 0.0386 & 0.0417 & 0.0341\\
Number of Explanations = 100 & 0.0244 & 0.0386 & 0.0417 & 0.0340\\
Number of Explanations = 200 & 0.0244 & 0.0385 & 0.0416 & 0.0340\\
\midrule
Temperature = 0 & 0.0247 & 0.0381 & 0.0420 & 0.0337\\
Temperature = 0.2 & 0.0245 & 0.0383 & 0.0417 & 0.0339\\
Temperature = 0.5 (default) & 0.0245 & 0.0386 & 0.0417 & 0.0341\\
Temperature = 0.8 & 0.0246 & 0.0386 & 0.0417 & 0.0340\\
Temperature = 1 & 0.0245 & 0.0386 & 0.0418 & 0.0341\\
\midrule
Batch Size = 10,000 tokens & 0.0245 & 0.0386 & 0.0417 & 0.0341\\
Batch Size = 50,000 tokens & 0.0245 & 0.0386 & 0.0417 & 0.0341\\
Batch Size = 100,000 tokens (default) & 0.0245 & 0.0386 & 0.0417 & 0.0341\\
\bottomrule
\end{tabular}}}
\end{table}

\begin{table}[htbp]
\centering
\caption{Comparison of the Prediction Performance for Robustness Check.}
\label{tbl:robpred}
{\tabfont\setlength{\tabcolsep}{4pt}
\fitwidth{%
\begin{tabular}{lcccc}
\toprule
\textbf{Datasets} & \multicolumn{2}{c}{\textbf{Bank Churn}} & \multicolumn{2}{c}{\textbf{HCT Survival}}\\
\cmidrule(lr){2-3}\cmidrule(lr){4-5}
\textbf{Evaluation Metrics} & \textbf{Accuracy} & \textbf{F1-Score} & \textbf{RMSE} & \textbf{MAE}\\
\midrule
Number of Explanations = 10 & 0.918 & 0.693 & 19.978 & 15.608\\
Number of Explanations = 20 & 0.920 & 0.696 & 19.876 & 15.523\\
Number of Explanations = 50 (default) & 0.921 & 0.698 & 19.848 & 15.502\\
Number of Explanations = 100 & 0.921 & 0.698 & 19.848 & 15.496\\
Number of Explanations = 200 & 0.920 & 0.696 & 19.846 & 15.493\\
\midrule
Temperature = 0 & 0.917 & 0.693 & 19.923 & 15.538\\
Temperature = 0.2 & 0.921 & 0.693 & 19.863 & 15.509\\
Temperature = 0.5 (default) & 0.921 & 0.698 & 19.848 & 15.502\\
Temperature = 0.8 & 0.921 & 0.698 & 19.842 & 15.506\\
Temperature = 1 & 0.919 & 0.696 & 19.850 & 15.511\\
\midrule
Batch Size = 10,000 tokens & 0.923 & 0.696 & 19.898 & 15.523\\
Batch Size = 50,000 tokens & 0.923 & 0.701 & 19.839 & 15.497\\
Batch Size = 100,000 tokens (default) & 0.921 & 0.698 & 19.848 & 15.502\\
\bottomrule
\end{tabular}}}
\end{table}

\section{An Industrial Application: Live-Streaming CTR Prediction}
\label{sec:kuailive}

As the previous section establishes the performance of RashomonLLM on two offline benchmarks, we now turn to a large-scale, real-world application to demonstrate its practical value at industrial scale. Specifically, we use interaction logs from \emph{KuaiLive} \citep{qu2025kuailive}, a dataset collected on Kuaishou's live-streaming platform, one of the largest live-streaming platforms in China. The dataset records 17,615,350 exposure events (4,909,515 clicks and 12,705,835 exposed-but-not-clicked impressions) involving 23,772 users and 452,621 streamers spanning May 4--25, 2025. This collection window postdates the March 2025 knowledge cutoff of the Qwen3-30B-A3B backbone (released April 2025) \citep{qwen3_2025}, so none of these interactions could have been memorized during pretraining, ruling out the data leakage concern. We focus on the click-through rate (CTR) prediction task, a core problem in platform recommendation \citep{zhou2018deepinterest}, where the goal is to predict whether the user clicks into the live room given a user--streamer exposure. This setting is considerably more demanding than the offline benchmarks, as the interaction log is at an industrial scale that far exceeds the LLM context window, and the prediction is driven heavily by dynamic behavioral signals, rather than only static features.

\subsection{Experimental Setup}

We instantiate RashomonLLM on an open-weight LLM backbone Qwen3-30B-A3B with our Batch LLM Learning mechanism (Section~\ref{sec:batch}), because the full interaction log greatly exceeds the context window. We adopt an open-weight backbone, rather than the proprietary GPT-4o used for the offline benchmarks, due to the fact that keeping proprietary interaction logs within the platform's environment is essential in industrial settings. Another practical consideration is cost: self-hosting an open-weight model such as Qwen3 is substantially cheaper than paying per-token API fees to a proprietary model like GPT-4o at industrial volumes. Importantly, this choice does not compromise performance, as the accuracy lead is not specific to any single LLM, according to Table~\ref{tbl:kuaibackbone} in Section~\ref{sec:ablation}. We utilize both the static and contextual user, streamer, and live-room attributes (e.g., demographics, device tier, streamer content niche, time-of-day), and the dynamic interaction features (historical user CTR, streamer CTR, content-category CTR, activity counts), each computed strictly from pre-exposure history to prevent temporal leakage, for the prediction task. Because the natural click rate on the platform is only $\sim$28\%, we train and evaluate on a class-balanced sample of the log ($\approx$9.82M records: all 4{,}909{,}515 clicks and an equal number of randomly sampled non-clicks), so that accuracy is a meaningful discriminative metric and learning is not biased toward the no-click majority. We compare against a comprehensive set of baselines spanning classical tabular predictors (Logistic Regression, Gradient-Boosted Trees (GBM), and a tuned LightGBM \citep{ke2017lightgbm}), a general-purpose tabular deep model (FT-Transformer \citep{gorishniy2021ft}), and specialized click-through rate prediction models deployed in the industry, including Factorization Machine \citep{rendle2010fm}, Wide \& Deep \citep{cheng2016widedeep}, DeepFM \citep{guo2017deepfm}, and DCNv2 \citep{wang2021dcnv2}, all trained on the identical feature set. We adopt an 80/20 train/test split, set the EPR stopping threshold to $\theta = 0.01$, and process the training log with Batch LLM Learning in sequential batches of 50{,}000 tokens, each fed through the context window in turn, as described in Section~\ref{sec:batch}. For explanation quality, we evaluate against the single-shot LLM post-hoc explainer, the Sparse Decision Tree baseline \citep{xin2022}, and a permutation-importance reference \citep{breiman2001rf} applied on top of DCNv2. The prompt templates for the three EPR agents on this CTR task are provided in Table~\ref{tbl:prompts}.

\subsection{Prediction Performance}

As reported in Table~\ref{tbl:kuaipred}, RashomonLLM significantly outperforms both strong classical baselines and the broader suite of specialized, industry-deployed CTR architectures. In particular, it achieves significant performance improvements over every single baseline model across all three evaluation metrics, while producing faithful natural-language explanations at the same time, as we will demonstrate in the next section. This is a notable result, as we illustrate that on a large-scale tabular prediction task, where gradient-boosted trees and dedicated CTR models are considered state of the art, RashomonLLM not only matches but exceeds them, confirming that the accuracy gains established on the offline benchmarks carry over to the industrial application.

\begin{table}[htbp]
\centering
\caption{CTR Prediction Performance on the KuaiLive Industrial Dataset. ***p$<$0.01.}
\label{tbl:kuaipred}
{\tabfont
\fitwidth{%
\begin{tabular}{lccc}
\toprule
\textbf{Model} & \textbf{Accuracy} & \textbf{F1} & \textbf{AUC}\\
\midrule
\textbf{RashomonLLM} & \textbf{0.771***} & \textbf{0.772***} & \textbf{0.842***}\\
Logistic Regression & 0.748 & 0.747 & 0.798\\
Factorization Machine & 0.749 & 0.748 & 0.811\\
Gradient-Boosted Trees & 0.752 & 0.752 & 0.821\\
LightGBM (tuned) & 0.756 & 0.755 & 0.824\\
Wide \& Deep & 0.758 & 0.757 & 0.826\\
FT-Transformer & 0.760 & 0.759 & 0.828\\
DeepFM & 0.761 & 0.760 & 0.831\\
DCNv2 & 0.761 & 0.759 & 0.833\\
\bottomrule
\end{tabular}}}
\end{table}

Because RashomonLLM uses a large language-model backbone, a natural concern is that this lead simply reflects the capacity of the backbone rather than our design. The ablation in Section~\ref{sec:ablation} rules this out. The same backbone in a prediction-only setting, without the explanation--prediction coupling, reaches only 0.650 accuracy, which is well below every specialized baseline in Table~\ref{tbl:kuaipred}, and an iteratively trained variant with an uninformative rationale of matched length still underperforms the specialized models. Only when prediction is coupled with a faithful self-explanation does accuracy rise to 0.771 and surpass them. The lead is therefore attributed to the fidelity of the explanation, consistent with Theorem~2, rather than to the raw capacity of the LLM backbone.

\subsection{Explanation Quality and Examples}

We next focus on the model comparison of explanation quality. In this large-scale study, we additionally report two faithfulness metrics \citep{deyoung2020}: \emph{Comprehensiveness}, the additional drop in performance when all features the explanation deems important are removed, and \emph{Sufficiency}, the drop in performance when only those important features are kept (so lower is better). On this classification task, all four faithfulness metrics are reported as changes in prediction accuracy ($\Delta$accuracy). As reported in Table~\ref{tbl:kuaiexpl}, RashomonLLM produces the most faithful explanations on KuaiLive, since deleting its top-ranked features induces the largest accuracy drop of any method, while its predictions are also the most robust when non-influential features are perturbed. Since it significantly outperforms the single-shot LLM post-hoc explainer, we confirm the explanation quality gain comes from our Rashomon/EPR design, rather than from merely prompting an LLM for explanations. Beyond outperforming dedicated XAI methods on these faithfulness metrics, RashomonLLM uniquely generates natural-language, per-instance rationales that the traditional feature-importance baselines cannot provide, which are illustrated in Table~\ref{tbl:kuaiexamples}, where each rationale is grounded in the specific feature values of the record and is directly readable by a non-technical stakeholder. They are contrastive and propensity-aware: a comparison of records 1 and 2, two users with extensive activity histories, shows that the model attributes record 1's predicted click primarily to the user's high personal click-through rate, and correctly withholds a click prediction for the equally active but low-propensity user in record 2. Additionally, the rationales lead to actionable insights, such as those on the matched content category and the time-of-day, that a platform can directly influence through ranking and scheduling.

\begin{table}[htbp]
\centering
\caption{Explanation Quality on the KuaiLive Dataset. ***p$<$0.01.}
\label{tbl:kuaiexpl}
{\tabfont\setlength{\tabcolsep}{4pt}
\fitwidth{%
\begin{tabular}{lcccc}
\toprule
\textbf{Method} & \textbf{Single Deletion ($\uparrow$)} & \textbf{Random.\ Check ($\downarrow$)} & \textbf{Compreh.\ ($\uparrow$)} & \textbf{Suffic.\ ($\downarrow$)}\\
\midrule
\textbf{RashomonLLM} & \textbf{0.273***} & \textbf{0.020***} & \textbf{0.315***} & \textbf{0.041***}\\
Sparse Decision Tree & 0.240 & 0.033 & 0.258 & 0.077\\
Permutation Importance & 0.244 & 0.028 & 0.271 & 0.069\\
LLM Post-hoc Explainer & 0.201 & 0.059 & 0.238 & 0.104\\
\bottomrule
\end{tabular}}}
\end{table}

\begin{table}[htbp]
\centering
\caption{Selected Per-Instance Explanations Produced by RashomonLLM for KuaiLive Test Users.}
\label{tbl:kuaiexamples}
{\tabfont
\sbox0{\begin{tabular}{p{0.29\linewidth} c p{0.45\linewidth}}
\toprule
\textbf{Records (key feature values)} & \textbf{Pred.} & \textbf{RashomonLLM explanation}\\
\midrule
\#1 user\_ctr 0.66, streamer\_ctr 0.38, cat\_ctr 0.29, hour 14; extensive watch history & Click & ``Because the user's high historical CTR (0.66) and the streamer's strong CTR (0.38), together with the user's extensive live-watching history, collectively indicate strong engagement potential.''\\
\addlinespace
\#2 user\_ctr 0.02, user\_n 5{,}949 (very active), streamer\_ctr 0.03, hour 12 & No click & ``Because the user's CTR (0.02) is very low and the streamer's CTR (0.03) is low; despite the user's heavy browsing activity, the low historical interaction rates indicate a low likelihood of a click.''\\
\addlinespace
\#3 user\_ctr 0.34, streamer\_ctr 0.53, hour 20 (peak) & Click & ``Because the user's high CTR (0.34) and the streamer's very high CTR (0.53), with exposure at a peak hour, all indicate strong engagement potential.''\\
\bottomrule
\end{tabular}}%
\usebox0\par\vspace{2pt}
\parbox{\wd0}{\footnotesize\raggedright \textit{Note.} Each rationale is produced by conditioning the learned, aggregated explanation on the individual record, rather than by re-running the EPR loop per user.\par}}
\end{table}

\subsection{Reflection Dynamics and Convergence}

\begin{figure}[htbp]
\centering
\includegraphics[width=0.7\linewidth]{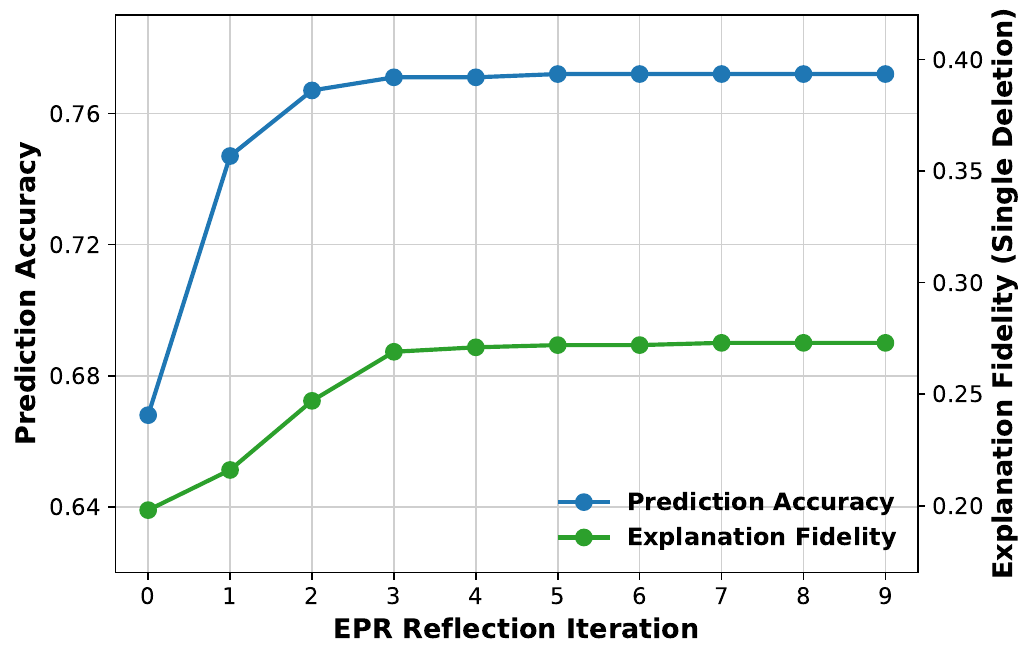}
\caption{Prediction Accuracy and Explanation Fidelity (Single-Deletion Faithfulness) Across EPR Reflection Iterations on KuaiLive, Co-Measured on the Same 1{,}000 Held-Out Records.}
\label{fig:complementarity}
\end{figure}

A distinctive feature of RashomonLLM is that the explanation itself is revised through the Reflection mechanism, which we illustrate in Table~\ref{tbl:kuaiepr}. The initial explanation conflates engagement with device and demographic attributes and already predicts moderately well (accuracy 0.668). The first reflection step, prompted by the systematic errors on the validation set, recognizes that historical click-through and watch-time signals drive clicks regardless of device price or content tag, and lifts accuracy to 0.747. Subsequent reflections sharpen the click-through rate thresholds with only diminishing gains (0.767 and 0.771). This trajectory is consistent with the boundary condition in our theory, as reflection is most valuable while the explanation still misallocates importance, and its benefit diminishes once the explanation has captured the dominant signal. We further visualize the full ten-iteration trajectory in Figure~\ref{fig:complementarity}, which shows that explanation quality improves along with prediction accuracy over the first few iterations and then converges quickly, consistent with the geometric contraction we verify next.

\begin{table}[htbp]
\centering
\caption{Progression of the RashomonLLM Explanation Across EPR Reflection Iterations on KuaiLive. Validation accuracy is measured on 1,000 held-out records.}
\label{tbl:kuaiepr}
{\tabfont
\fitwidth{%
\begin{tabular}{c c p{0.48\linewidth} p{0.4\linewidth}}
\toprule
\textbf{Iter.} & \textbf{Val. Acc.} & \textbf{Click-pattern description (excerpt)} & \textbf{What reflection changed}\\
\midrule
0 & 0.668 & Predicts a click when user CTR $\geq$0.15 and device price is high ($\geq$5000), or for older users on premium devices. & Entangles genuine engagement signals with device and demographic attributes.\\
\addlinespace
1 & 0.747 & A click follows from high user CTR ($\geq$0.15) or high cumulative watch time, regardless of device price, and from streamer popularity regardless of content tag. & Drops the spurious device-price requirement; largest single-step gain.\\
\addlinespace
2 & 0.767 & High user CTR ($\geq$0.3) becomes a dominant signal even with low watch time, with a user--streamer engagement interaction governing the no-click cases. & Refines thresholds and adds the user--streamer engagement interaction for no-click cases.\\
\addlinespace
3 & 0.771 & CTR-dominant rule with a tightened no-click condition (low user CTR and low streamer CTR). & Consolidates the CTR-dominant rule; marginal gain tapers as the description saturates.\\
\bottomrule
\end{tabular}}}
\end{table}

\begin{table}[htbp]
\centering
\caption{Contraction Diagnostic for the EPR Update on KuaiLive.}
\label{tbl:kuaicontract}
{\tabfont
\fitwidth{%
\begin{tabular}{ccc}
\toprule
\textbf{Step ($t \to t{+}1$)} & $\lVert e_{t+1}-e_{t}\rVert_{2}$ & \textbf{Ratio to previous step}\\
\midrule
$0 \to 1$ & 0.79 & ---\\
$1 \to 2$ & 0.31 & 0.39\\
$2 \to 3$ & 0.11 & 0.35\\
$3 \to 4$ & 0.04 & 0.36\\
\bottomrule
\end{tabular}}}

{\footnotesize\raggedright \textit{Note.} All displacements beyond the fourth iteration are negligible and are omitted, as the explanation has effectively converged.\par}
\end{table}

The same trajectory also lets us empirically test the contraction property in Assumption~3. Specifically, we take the embedding map $\Phi(\cdot)$ to be the backbone LLM's own representation of the explanation text, and measure the per-iteration displacement $\lVert e_{t+1}-e_{t}\rVert_{2} = \lVert \Phi(E_{t+1})-\Phi(E_{t})\rVert_{2}$, which Assumption~3 predicts should decay geometrically. Table~\ref{tbl:kuaicontract} confirms this, where the displacement shrinks by a roughly constant factor (ratio $k \approx 0.4 < 1$), providing direct empirical support for the assumption behind our convergence results.

\subsection{Ablation Studies}\label{sec:ablation}

To establish that RashomonLLM's performance stems from our specific design choices, we conduct four groups of ablation studies in this section, where we construct a series of variant models for each group, which are trained and evaluated on the same balanced KuaiLive split used for our main results. We summarize the four ablations and their key findings in Table~\ref{tbl:kuaiablsummary}.

\begin{table}[htbp]
\centering
\caption{Overview of the Ablation Studies.}
\label{tbl:kuaiablsummary}
{\tabfont\setstretch{1}
\fitwidth{%
\begin{tabular}{>{\RaggedRight}p{0.18\linewidth} >{\RaggedRight}p{0.25\linewidth} >{\RaggedRight}p{0.48\linewidth}}
\toprule
\textbf{Ablation} & \textbf{What it probes} & \textbf{Key finding}\\
\midrule
Feature Groups & Source of absolute accuracy & Dynamic features are decisive for CTR prediction.\\
\addlinespace
Method Components & Source of performance gain & Self-explanation supplies the gain---driven by explanation fidelity, not added text---while reflection adds a further gain on top of the fixed-explanation variant.\\
\addlinespace
Explanation Size & Number of explanations & Accuracy/faithfulness rise with $k$ and saturate near 50.\\
\addlinespace
Backbone \& Size & Backbone/size dependence & The lead holds across LLM backbones and sizes.\\
\bottomrule
\end{tabular}}}
\end{table}

\stat{Feature Groups.} We construct three variants that differ only in the features exposed to every model: the \emph{Static-only} variant uses only the static user, streamer, and live-room attributes; the \emph{$+$\,User \& Streamer CTR} variant augments the static set with the two historical CTR features for the user and the streamer; and the \emph{Full} variant additionally includes the content-category CTR and the user and streamer activity counts. As Table~\ref{tbl:kuaiablate} shows, with static attributes alone, none of these models perform well, whereas the full dynamic feature set lifts every model by roughly seven accuracy points on average, of which the two historical user and streamer CTRs alone account for about five, indicating the importance of behavioral features in the CTR prediction task.

\begin{table}[htbp]
\centering
\caption{Feature-Group Ablation on KuaiLive (Balanced-Test Accuracy).}
\label{tbl:kuaiablate}
{\tabfont
\fitwidth{%
\begin{tabular}{lccc}
\toprule
\textbf{Feature set (variant)} & \textbf{RashomonLLM} & \textbf{GBM} & \textbf{LogReg}\\
\midrule
Static attributes only & 0.703 & 0.694 & 0.669 \\
\quad $+$ user \& streamer CTR & 0.752 & 0.735 & 0.728 \\
Full ($+$ category CTR \& activity counts) & 0.771 & 0.752 & 0.748\\
\bottomrule
\end{tabular}}}
\end{table}

\stat{Method Components.} We decompose RashomonLLM along its two design axes, namely one-shot in-context use versus iterative training, and the presence of an authentic self-explanation, resulting in four variants: \emph{One-shot, prediction-only} runs the Prediction Agent alone on raw features, with no explanation or reflection; \emph{One-shot $+$ reflection} adds the Reflection Agent but only runs it once; \emph{Iterative, without reflection} moves to iterative training but omits the reflection step with a fixed explanation prompt; and \emph{Iterative $+$ placebo explanation} retains the iterative training but replaces the authentic rationale with an uninformative one of matched length, i.e., a randomly drawn, instance-irrelevant template, which isolates whether the gain reflects explanation fidelity or merely the generation of additional text. As we observe in Table~\ref{tbl:kuaicomp}, the placebo variant falls well short of the authentic self-explanation, so the gain comes from explanation fidelity rather than the training procedure or the added text. The raw LLM prediction alone also falls below every specialized baseline, so the LLM is not inherently superior to baseline models. While adding reflection and moving to iterative training help with the performance, only the full, reflection-refined self-explanation attains the highest accuracy. We therefore confirm that our method's performance gain comes from reflecting on and articulating a faithful explanation.

\begin{table}[htbp]
\centering
\caption{Method-Component Ablation on KuaiLive.}
\label{tbl:kuaicomp}
{\tabfont
\fitwidth{%
\begin{tabular}{lc}
\toprule
\textbf{Configuration (variant)} & \textbf{Accuracy}\\
\midrule
One-shot, prediction-only & 0.650\\
One-shot $+$ reflection & 0.689\\
Iterative, without reflection (fixed explanation prompt) & 0.751\\
Iterative $+$ placebo explanation & 0.711\\
Full model (RashomonLLM) & 0.771\\
\bottomrule
\end{tabular}}}
\end{table}

\stat{Explanation Size.} We vary the number of retained explanations $k$ aggregated into the final explanation. As Table~\ref{tbl:kuaiksweep} shows, both accuracy and explanation faithfulness increase with $k$ and saturate near $k=50$, beyond which additional explanations add negligible value.

\begin{table}[htbp]
\centering
\caption{Explanation Size Ablation on KuaiLive.}
\label{tbl:kuaiksweep}
{\tabfont
\fitwidth{%
\begin{tabular}{lcc}
\toprule
\textbf{Set size $k$} & \textbf{Accuracy} & \textbf{Single Deletion}\\
\midrule
1   & 0.755 & 0.251\\
10  & 0.765 & 0.269\\
25  & 0.768 & 0.272\\
50  & 0.771 & 0.273\\
100 & 0.771 & 0.273\\
\bottomrule
\end{tabular}}}
\end{table}

\stat{Backbone and Capacity.} To test whether the result is specific to our chosen LLM backbone, we re-run RashomonLLM on alternative open-weight models spanning an order of magnitude in scale and two model families: Llama-3.1 and Qwen3. As Table~\ref{tbl:kuaibackbone} shows, the performance gain is neither backbone- nor scale-specific, as the 30B or 235B model adds only marginal value. Capacity is thus not the bottleneck, consistent with the predictive signal being concentrated in CTR features.

\begin{table}[htbp]
\centering
\caption{Backbone Ablation on KuaiLive (Balanced-Test Accuracy).}
\label{tbl:kuaibackbone}
{\tabfont
\fitwidth{%
\begin{tabular}{llc}
\toprule
\textbf{Backbone (variant)} & \textbf{Size} & \textbf{Accuracy}\\
\midrule
Llama-3.1-8B-Instruct & 8B & 0.767\\
Qwen3-8B-Instruct & 8B & 0.768\\
Qwen3-30B-A3B (default) & 30B / 3B active & 0.771\\
Qwen3-235B-A22B-Instruct & 235B / 22B active & 0.773\\
\bottomrule
\end{tabular}}}
\end{table}

\stat{What the ablations show.} As summarized in Table~\ref{tbl:kuaiablsummary}, RashomonLLM's absolute accuracy comes from the CTR features, but its lead over strong baselines is created by the method's components, and, most importantly, by equipping the learned model to explain itself. Therefore, a faithful natural-language rationale does not trade off against accuracy but improves it, so explanation and prediction are complementary, as Theorem~2 predicts.

\subsection{Robustness Checks}\label{sec:kuairobustness}

In robustness checks, we vary the evaluation conditions, re-running the learning procedure under each condition where the check requires it (e.g., the random-seed and time-split checks), studying whether the reported findings survive perturbations that commonly undermine LLM-based studies. Table~\ref{tbl:kuairobsummary} previews the eight perturbation checks and their outcomes.

\begin{table}[htbp]
\centering
\caption{Overview of the Robustness Checks.}
\label{tbl:kuairobsummary}
{\tabfont\setstretch{1}
\fitwidth{%
\begin{tabular}{>{\RaggedRight}p{0.17\linewidth} >{\RaggedRight}p{0.32\linewidth} >{\RaggedRight}p{0.48\linewidth}}
\toprule
\textbf{Check} & \textbf{What it probes} & \textbf{Key finding}\\
\midrule
LLM Temperature & Decoding stochasticity & Predictions are near-invariant to decoding randomness.\\
\addlinespace
Feature Order & Reliance on positional cues & Predictions are invariant to feature ordering.\\
\addlinespace
Prior Shift & External validity under the true prior & Ranking quality is preserved under the true click prior.\\
\addlinespace
Random Seed & Initialization/shuffle impact & The lead exceeds run-to-run seed variability.\\
\addlinespace
Time-based Split & Temporal generalization & The advantage persists under time-based split.\\
\addlinespace
Cold-start & No-history degradation & Accuracy degrades slightly for cold-start users.\\
\addlinespace
Calibration & Probability usability & Probabilities are useful for ranking and bidding.\\
\addlinespace
Leakage Placebo & Temporal leakage in CTR features & No temporal leakage exists.\\
\bottomrule
\end{tabular}}}
\end{table}

\stat{LLM Temperature.} We change the temperature to $T=1.0$, versus $T=0.5$ in our main analysis. Results in Table~\ref{tbl:kuairobust} show that there is no significant performance change related to the temperature. 

\stat{Feature Order.} We randomly permute the order of features in the prompt and the performance is essentially unchanged in Table~\ref{tbl:kuairobust}, so the model depends on feature values rather than order. 

\stat{Prior Shift.} Since the model is trained on a balanced sample whereas the platform's true click rate is only $\sim$28\%, we resample the test set to that natural prior. Table~\ref{tbl:kuairobust} confirms that ranking quality transfers to the natural operating distribution with negligible changes in AUC. 

\stat{Random Seed.} We repeat the entire Batch LLM Learning five times with different random seeds and report the mean and standard deviation. With an accuracy of 0.770 and a standard deviation of 0.004 reported in Table~\ref{tbl:kuairobust}, we show that our performance gain is not an artifact of a specific seed. 

\stat{Time-based Split.} Because recommendation is inherently temporal, we replace the random split with a time-based one, where we train on the earlier portion of the interaction log and test on the later portion. Table~\ref{tbl:kuairobust} confirms that our advantage persists under this alternative splitting strategy. 

\stat{Cold-start.} We study the performance of our model on users and streamers with no prior history, where we only observe marginal performance degradation reported in Table~\ref{tbl:kuairobust}. 

\stat{Calibration.} We measure the expected calibration error (ECE) of the predicted click probabilities, where RashomonLLM's ECE is 0.031 versus 0.042 for the gradient-boosted tree (Table~\ref{tbl:kuairobust}), so its predicted probabilities are better calibrated and usable for ranking and bidding.

\stat{Leakage Placebo.} To test for temporal leakage in CTR features, we recompute them from a randomly time-shuffled history and Table~\ref{tbl:kuairobust} shows that the time-shuffled placebo collapses accuracy, confirming that the CTR features carry genuinely historical, non-leaked signal.

\begin{table}[htbp]
\centering
\caption{Robustness Checks on KuaiLive.}
\label{tbl:kuairobust}
{\tabfont
\fitwidth{%
\begin{tabular}{lccc}
\toprule
\textbf{Setting (variant)} & \textbf{Accuracy} & \textbf{F1} & \textbf{AUC}\\
\midrule
RashomonLLM & 0.771 & 0.772 & 0.842\\
LLM temperature $T{=}1.0$ & 0.766 & 0.767 & 0.834\\
Feature-order permutation & 0.769 & 0.770 & 0.841\\
Natural prior ($\sim$28\% CTR) & 0.786 & 0.717 & 0.833\\
Random seeds (5 runs) & 0.770 & 0.772 & 0.841 \\
Time-based split & 0.767 & 0.767 & 0.839 \\
Cold-start users (no history) & 0.753 & 0.756 & 0.821 \\
Cold-start streamers (no history) & 0.761 & 0.763 & 0.829 \\
Time-shuffled placebo & 0.658 & 0.671 & 0.756 \\
\midrule
\multicolumn{4}{p{0.62\linewidth}}{Probability calibration: expected calibration error (ECE) is $0.031$ for RashomonLLM versus $0.042$ for the gradient-boosted tree.}\\
\bottomrule
\end{tabular}}}
\end{table}

\stat{What the robustness checks show.} As summarized in Table~\ref{tbl:kuairobsummary}, our main findings hold across all robustness checks, as predictions are insensitive to temperature, feature ordering, random seed, splitting strategy, and click prior. The model degrades slightly in cold-start subgroups and significantly with time-shuffled placebo. RashomonLLM therefore delivers significant performance advantages with well-calibrated, faithful explanations.

\stat{Pretraining contamination.} A recurring concern for LLM-based predictors is that the backbone LLM may have memorized the evaluation data during pretraining. This, however, is an implausible account of our results due to the following reasons. First, the KuaiLive interaction logs were not public until the dataset's release \citep{qu2025kuailive}, so regardless of any backbone's knowledge cutoff they could not have entered its pretraining corpus, and our main results cannot be contaminated. Second, the predictive signal in KuaiLive resides overwhelmingly in dynamic CTR features that cannot appear in a pretraining corpus, and the model predominantly exploits these private signals rather than recalled facts. Finally, the placebo-rationale experiment confirms that RashomonLLM reasons from the input features, rather than recalling memorized outcomes.

\subsection{Practical Deployment Considerations}\label{sec:deployment}

We now discuss several considerations for stakeholders deploying RashomonLLM in practice.

\stat{Cost and economic gain.} Table~\ref{tbl:kuaicost} benchmarks the end-to-end cost of RashomonLLM against Gradient-Boosted Trees, a strong baseline popular in industrial practice \citep{grinsztajn2022}. RashomonLLM is more expensive to operate, but its training cost stays modest on a single 8$\times$H200 node with a throughput-amortized serving cost of \$0.004 per thousand predictions, well within the budget of a platform already operating large-scale recommendation-serving infrastructure \citep{gupta2020}. Meanwhile, RashomonLLM's accuracy advantage of 1.9 percentage points over Gradient-Boosted Trees yields about 19 additional correct decisions per thousand records. We value these using Kuaishou's FY2025 online-marketing revenue of RMB~208 per daily active user and the roughly 25{,}000 monetizable exposures a user receives per year \citep{kuaishou2025}. At that valuation, each additional correct decision is worth RMB~0.008 $\approx$US\$0.001, an economic gain on the order of US\$0.02 per thousand predictions, several times the serving cost.

\begin{table}[htbp]
\centering
\caption{Deployment Cost--Benefit Comparison on KuaiLive, per Thousand Predictions.}
\label{tbl:kuaicost}
{\tabfont\setlength{\tabcolsep}{4pt}
\fitwidth{%
\begin{tabular}{lccccc}
\toprule
\textbf{Model} & \textbf{Training (per iteration)} & \textbf{Serving latency} & \textbf{Cost (\$/1k)} & \textbf{Accuracy} & \textbf{Econ.\ gain (\$/1k)}\\
\midrule
Gradient-Boosted Trees & $<$1 min & $\sim$1 ms & $<$0.001 & 0.752 & ref.\\
RashomonLLM & $\sim$20 min & $\sim$5 ms & 0.004 & 0.771 & 0.02\\
\bottomrule
\end{tabular}}}
\end{table}

\stat{Latency and serving pattern.} We observe from Table~\ref{tbl:kuaicost} that serving latency is low for both models: roughly 1~ms per record for the gradient-boosted tree and 5~ms for RashomonLLM. This gap is negligible in practice \citep{gupta2020}, and RashomonLLM's additional cost concentrates on the one-time training side rather than real-time serving.

\stat{Privacy and data governance.} In a deployment scenario, because the backbone LLM is open-weight and can be adapted in-house, proprietary interaction logs need never leave the firm's environment, and the entire learning and inference pipeline would remain auditable within the organization, thereby addressing the privacy and compliance concerns in industrial settings.

\stat{Model refresh.} Because user and content distributions continually drift on a live platform, a deployed model must be refreshed periodically. The time-based split evaluation (Table~\ref{tbl:kuairobust}) shows that RashomonLLM generalizes forward in time and can be retrained on a scheduled cadence, such as daily or weekly batch updates. Since each adaptation is a single lightweight Batch LLM Learning run, such refreshes are inexpensive to operate.

\stat{Cold-start handling.} For users or streamers with no interaction history, our model still exploits static attributes to provide relatively good performance, as shown in Table~\ref{tbl:kuairobust}. In deployment, this leads to a fallback path, where we route these cases to a dedicated cold-start or exploration mechanism until enough interaction accumulates for the full RashomonLLM model to take over.

\subsection{Practical Implications}

Our study yields several implications for the governance of digital recommendation platforms.

\emph{First, the recommendation model should be both accurate and self-explaining.} RashomonLLM exceeds specialized recommendation models while producing its own rationales, which lowers the engineering and maintenance overhead and, more importantly, resolves the accuracy--explainability trade-off that typically becomes a burden for deploying XAI methods.

\emph{Second, natural-language explanations make model decisions consumable by non-technical stakeholders.} Product managers, creator-relations teams, content reviewers, and regulators can read why a particular exposure was or was not predicted to convert, without specialized training in feature-attribution methods. This directly supports accountability requirements and the ``right to explanation'', and it turns the model from a black-box scorer into a communicable decision aid, which is an essential property for the organizational adoption of AI.

\emph{Third, the explanations reveal managerial and actionable insights.} The dominance of dynamic CTR features over static demographics indicates that platforms should prioritize investment in real-time behavioral feature pipelines rather than richer profile data. The contrast between records~\#1 and~\#2 in Table~\ref{tbl:kuaiexamples} cautions against a common targeting heuristic: a highly active user is not necessarily a likely clicker, and the engagement volume should not be conflated with click propensity.

\emph{Fourth, our proposed solution is economical and portable.} It runs on an open-weight backbone at modest serving cost with user logs kept in-house (Section~\ref{sec:deployment}), and the core model architecture transfers easily to other business tasks, such as purchase and churn prediction, making interpretable prediction accessible to firms without large dedicated XAI teams.

\emph{Finally, boundary conditions are important.} The reflection mechanism adds predictive value when the base predictor still has room for improvement, but contributes little once the predictor is saturated, so practitioners should expect its largest gains in sparse or long-tail cases. Likewise, while Batch LLM Learning and inference scale to millions of interactions in our study, further engineering might be required for deployment at a larger scale, which we discuss further in Section~\ref{sec:scope}.

\section{Discussion}\label{sec:discussion}

In this section, we discuss theoretical and practical contributions, as well as boundary conditions.

\subsection{Theoretical Contributions}\label{sec:contributions}

In terms of design-science research \citep{gregorhevner2013}, ours is an \emph{improvement} contribution, where we propose a new class of solution (i.e., coupling explanation with prediction) to a problem the XAI literature has long treated as an inherent accuracy--explainability trade-off \citep{gunning2019}. What makes this contribution general, rather than specific to our artifact, is the formal theoretical guarantees that we present in this paper: the paradigm-level results (Theorems~1--3) hold independently of any particular model as long as the assumptions are satisfied, while Theorem~4 characterizes the convergence and coverage of the EPR workflow itself. In particular, Theorem~2 identifies the mechanism that explains why coupling explanation and prediction dissolves the trade-off, extending the IS tradition on system-generated explanations \citep{gregor1999} from a means of supporting user reasoning to a driver of predictive performance. We abstract this account into three prescriptive design principles (Section~\ref{sec:orgtheory}) that operationalize the development of other agentic XAI systems, as we specify what such systems should produce, how they should revise it, and how their components should be organized, so that the design knowledge transfers beyond RashomonLLM itself.

\subsection{Implications for Research and Practice}\label{sec:implications}

For practitioners, our results suggest that the value of an explanation should be assessed by its predictive usefulness, not only its post-hoc plausibility, and that explanation generation and prediction are best studied as a single coupled, iterative process rather than as separate stages. Our RashomonLLM offers an interpretable prediction model that delivers superior accuracy while producing faithful explanations that are directly consumable by the human and organizational processes, as an analyst can inspect, contest, or override a recommendation on the basis of a stated reason rather than a black-box score. The value of these capabilities is also translated into monetary terms in our KuaiLive study (Table~\ref{tbl:kuaicost}). We deliberately refrain from claiming any effect on users' \emph{perceived} trust or decision quality, since it is distinct from explanation fidelity and governed by plausibility and perception biases, and can even be inflated by unfaithful explanations \citep{lakkaraju2020}. What our results do establish are the prerequisites for that value to be realized: the explanations are faithful to the model, and the model they explain is more, not less, accurate. More broadly, our proposed approach enables organizations to pursue predictive performance and stakeholder trust simultaneously.

\subsection{Boundary Conditions}
\label{sec:scope}

We now discuss the conditions under which the method is and is not appropriate, so as to guide its responsible adoption in industrial practices.

\stat{Data modality and feature interpretability.} RashomonLLM is designed for tabular, structured inputs whose features carry meaningful semantics, and its explanations are only as interpretable as the underlying features. In the KuaiLive study, we deliberately exclude opaque inputs such as anonymized one-hot codes and learned title embeddings, because no faithful natural-language rationale can be attached to dimensions that lack semantics. Therefore, applications dominated by raw, unstructured signals, such as images, audio, or free text, fall outside our present scope and would require either encoding those signals into interpretable intermediate features or extending the LLM backbone with multimodal capability, a direction we leave to future work.

\stat{Task structure and feature dimensionality.} The method is strongest when the decision is governed by patterns that can be expressed compactly in language, i.e., interactions among a moderate number of features, as in churn and click prediction. When the predictive signal is instead distributed across hundreds of weakly informative features with high-order interactions, a concise natural-language description may not be able to capture the full decision surface, and a specialized model that is not constrained to remain verbalizable may hold an advantage on prediction accuracy.

\stat{Data regime and the value of reflection.} The benefit of the EPR reflection loop is regime-dependent. When the predictor has room for improvement, such as in cases of small samples, long-tail users or items, or weak initial features, reflection significantly improves both the explanation quality and prediction accuracy, as we show in Table~\ref{tbl:kuaiepr}. Once the predictor is saturated, i.e., trained on abundant data with strong features, further reflection yields diminishing returns and plateaus.

\stat{Faithfulness needs to be verified.} Finally, because large language models can produce fluent but post-hoc rationalizations, the faithfulness of RashomonLLM's explanations cannot be taken for granted. We therefore validate it with model-agnostic single-deletion and randomization-check tests against ground-truth and XAI references, both on the offline benchmarks (Table~\ref{tbl:explquality}) and on KuaiLive (Table~\ref{tbl:kuaiexpl}), and we recommend that any deployment include analogous faithfulness checks. Because a confident, well-formed rationale can nonetheless accompany an incorrect prediction, explanations should be evaluated for fidelity rather than mere plausibility.

\subsection{Limitations and Future Research}\label{sec:limitations}

\stat{Ethical use of sensitive attributes.} Several of our datasets contain demographic attributes, such as gender and age on KuaiLive, and age and race on the HCT survival task, and RashomonLLM's explanations may reference them. We note that our method is attribute-agnostic, as protected features are not required by the framework and can be excluded or masked without altering the workflow. We plan to study the impact of doing so on RashomonLLM in our future work.

\stat{Beyond tabular data.} We restrict the scope of this paper to tabular, structured inputs with semantically meaningful features. Extending the framework to unstructured data, such as text and images, either by encoding them into interpretable intermediate features or by leveraging multimodal backbones, will be an important next step to further enhance its practicality.

\stat{Live experiment.} A live online A/B experiment, together with improving the efficiency of training and inference for real-time, large-scale serving, remains an important direction for future work.

\stat{Learning from human explanations.} Connecting the LLM-generated explanations to small-sample learning with human-provided ground-truth explanations could further improve fidelity and enable human-in-the-loop refinement of our framework.

\stat{Ascending to causal explanation (Level~4).} Throughout the paper, we have been careful not to claim that RashomonLLM's rationales recover the true causal mechanism (Level~4). Closing that gap is the most consequential extension, and it requires changing both the evidence and the criterion of success. First, validation must shift from \emph{held-out predictive fidelity} to \emph{interventional consistency}. Second, this demands interventional data, such as a randomized-exposure A/B experiment, together with an explicit causal graph and its identification assumptions \citep{peters2016}. We regard this ascent from explaining the model to explaining the mechanism as the ultimate goal of the research pipeline built from this work.

\section{Conclusion}\label{sec:conclusion}

XAI is an important research area with significant implications for business applications, yet existing methods suffer from critical limitations that yield unreliable explanations and miss the opportunity for explanations to improve model performance. This paper addresses these limitations on three fronts. \emph{Theoretically}, we establish that no single explanation is robust to distribution shift, that an explanation's fidelity bounds the prediction error of the model it guides, and that the set of faithful-and-useful explanations---the Rashomon Explanation set---is non-empty and reachable. \emph{Methodologically}, we propose RashomonLLM, an Explanation--Prediction--Reflection agentic workflow with a Batch LLM Learning extension for industrial-scale data, and prove that it, under explicit conditions, converges within each run and recovers the full Rashomon Explanation set. These design choices are grounded in the organizational theories of sensemaking, double-loop learning, and complementarity, which together yield three design principles for agentic XAI. \emph{Empirically}, across two Kaggle benchmarks and a large-scale CTR prediction task at a live-streaming platform, RashomonLLM significantly outperforms state-of-the-art prediction and XAI baselines on both accuracy and explanation faithfulness; most notably, equipping the model to explain itself \emph{improves its predictions}, which offers direct evidence that explanation and accuracy are complementary rather than competing. By making them mutually reinforcing, RashomonLLM enables organizations to advance business performance and lay the groundwork for consumer trust at the same time.

\begin{singlespace}
\renewcommand{\bibliographytypesize}{\normalsize}
\bibliographystyle{apacite}
\bibliography{references}
\end{singlespace}

\end{document}